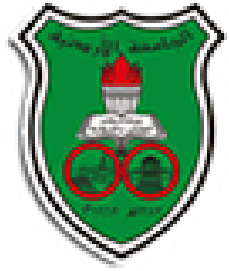

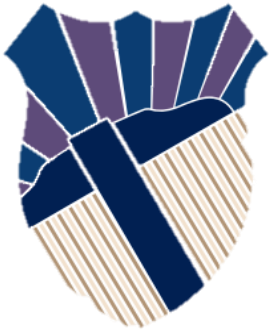
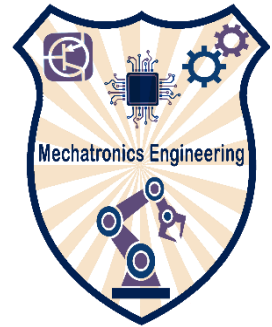

**School of Engineering**     **Department of Mechatronics Engineering**

# Bachelor of Science in Mechatronics Engineering
## Senior Design Graduation Project Report

## Design of Arabic Sign Language Recognition Model

**Report by**
Muhammad Al-Barham
Ahmad Jamal

**Supervisor**
Dr. Musa Al-Yaman

Date
26/05/2021

# ABSTRACT

Deaf people are using sign language for communication, and it is a combination of gestures, movements, postures, and facial expressions that correspond to alphabets and words in spoken languages. The proposed Arabic sign language recognition model helps deaf and hard hearing people communicate effectively with ordinary people.

The recognition has four stages of converting the alphabet into letters as follows: Image Loading stage, which loads the images of Arabic sign language alphabets that were used later to train and test the model, a pre-processing stage which applies image processing techniques such as normalization, Image augmentation, resizing, and filtering to extract the features which are necessary to accomplish the recognition perfectly, a training stage which is achieved by deep learning techniques like CNN, a testing stage which demonstrates how effectively the model performs for images did not see it before, and the model was built and tested mainly using PyTorch library.

The model is tested on ArASL2018, consisting of 54,000 images for 32 alphabet signs gathered from 40 signers, and the dataset has two sets: training dataset and testing dataset. We had to ensure that the system is reliable in terms of accuracy, time, and flexibility of use explained in detail in this report. Finally, the future work will be a model that converts Arabic sign language into Arabic text.

.

# TABLE OF CONTENTS

# LIST OF FIGURES

# LIST OF TABLES

# GLOSSARY

| ABBREVIATION | DESCRIPTION |
|---|---|
| ArSL | Arabic Sign Language |
| ANN | Artificial Neural Network |
| CNN | Convolution Neural Network |
| RNN | Recurrent Neural Network |
| ANFIS | Adaptive Neuro-Fuzzy Inference System |
| KNN | K Nearest Neighbour |
| SVM | Support Vector Machine |
| MLP | Multilayer Perceptron |
| GMM | Gaussian Mixture Model |
| LDA | Linear Discriminant Analysis |
| HOG | Histograms of Oriented Gradients |
| EHD | Edge Histogram Descriptor |
| DWT | Discrete Wavelet Texture |
| LBP | Local Binary Pattern |
| GLCM | Grey-Level Co-occurrence Matrix |
| BLEU | Bilingual Evaluation Understudy |
| TER | Translation Error Rate |
| ReLU | Rectified Linear Unit |
| SeLU | Scaled Exponential Linear Unit |
| AFOD | Arab Federation of the Deaf |
| ANFIS | Adaptive Neuro Fuzzy Inference |
| ArSLAT | Arabic Sign Language Automatic Translator |
| TPU | Tensor Processing Unit |
| ASICs | Application-Specific Integrated Circuits |
| CUDA | Compute Unified Device Architecture |
| SM | Streaming Multiprocessor |
| SGD | Stochastic Gradient Descent |

# Chapter 1 INTRODUCTION

## 1.1 Background

According to the World Health Organization (WHO), around 466 million people with hearing loss issues, and 34 million of them are children. It is claimed that by 2050 over 900 million people will have suffering hearing loss [1]. Hard hearing people can hear up to a specific limited degree and unobvious by a hearing aid. In contrast, deaf people cannot listen entirely due to head trauma, noise exposure, disease, or genetic condition [2].

Sign language is the means of communication between the deaf themselves and with ordinary people, and every country has its own language. One of these languages is ArSL, used in the Arabic regions; it was formally introduced in 2001 by the Arab Federation of the Deaf (AFOD) [3]. Sign Language depends on hand movements and gestures to accomplish what you want. There are various dialects of ArSL that differ from one country to another; it comprises 28 characters that the different dialects agree on them. Signs of Arabic alphabets is shown in Figure 1-1.

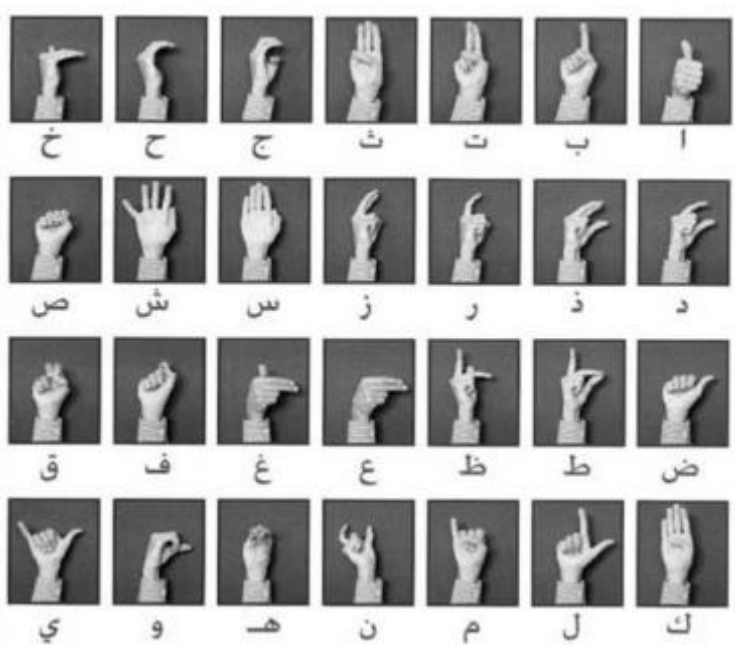

**Figure 1-1: Signs of Arabic alphabets** [4].

## 1.2 Problem Definition

The communication gap between ordinary people and deaf people is enormous, and we want to make it tiny, but it is a long road so, the best way is to study the subject from scratch and absorb the basis of it. As an initial point, we should learn in detail the signs of Arabic Alphabets to reduce sign language learners' obstacles, but the matter is not simple for all learners. Many of them will be confused when they learn a new field and may find this problematic. For this reason, we intend to build a model that recognizes the alphabet sign from Arabic Sign Language Speakers and then interprets it into text.

## 1.3 Literature Review

In [5], Omar Al-Jarrah and Alaa Halawani used a collection of ANFIS networks. Each network is trained to recognize one gesture. The system used images of bare hands, which allows the user to make the interaction more natural. The subtractive clustering algorithm and the least-squares estimator are used to identify the fuzzy inference system, and the training is achieved using the hybrid learning algorithm. The achieved accuracy is 93.55% that resulted from recognizing the 30 Arabic Manual alphabets.

In [6], Khaled Assaleh and M. Al-Rousan used Polynomial Classifiers to recognize the Arabic Sign Language Alphabet. It had seen that the Polynomial classifier has several advantages compared with ANFIS-based classification when they work on the same data. The data had been collected from deaf people and using the same corresponding feature set. The data collected by coloured Marked Glove-based systems. Polynomial Classifiers showed Significant results over ANFIS based on misclassified data patterns. Specifically, it has a 36% reduction when the methods were evaluated on the training data and a 57% reduction when the systems were assessed on the test data.

In [7], the author split the process into three stages; a data collection stage, a feature extraction stage, and a recognition stage using Hidden Markov Model (HMM). A vision-based methodology is used to collect the data, and then we need to prepare the data to absorb the necessary features to classify it using HMM. The collected data were 4500 signs from 15 samples with 300 signs for the single signer, 11 samples were taken for the training set; the accuracy obtained from the experiments is 88.73%.

This paper [8] shows an automated method for the translation of Arabic Sign Language alphabets. Its data had been collected using images of bare hands. The Experiments showed that the ArSLAT, Arabic Sign Language Automatic Translator, had an accuracy of 91.3% with 30 Arabic Alphabets.

In [9], the authors presented the stages they used to achieve recognition: skin detection, background exclusion, face and hands extraction, feature extraction, and classification using Hidden Markov Model (HMM). The dataset consists of 29 alphabet Arabic letters and numbers from 0to nine with different brightness. They used 253 training images and 104 testing images with 640×480 pixels. The recognition system is tested when dividing the handshape's rectangle surrounding it into 4, 9, 16, and 25 zones. At 16 zones, the recognition rate with 19 states reaches 100%, while at 4 and 9 zones cannot match 100%.

In [10], the author explained the nature of the dataset. It is images for the positive samples with the hand sign in different scales, different illumination in the complex background for each hand posture, and the negative samples images from the Internet which do not contain hand posture. Figure 1-2 shows the stages of translation from Natural language into Sign language.

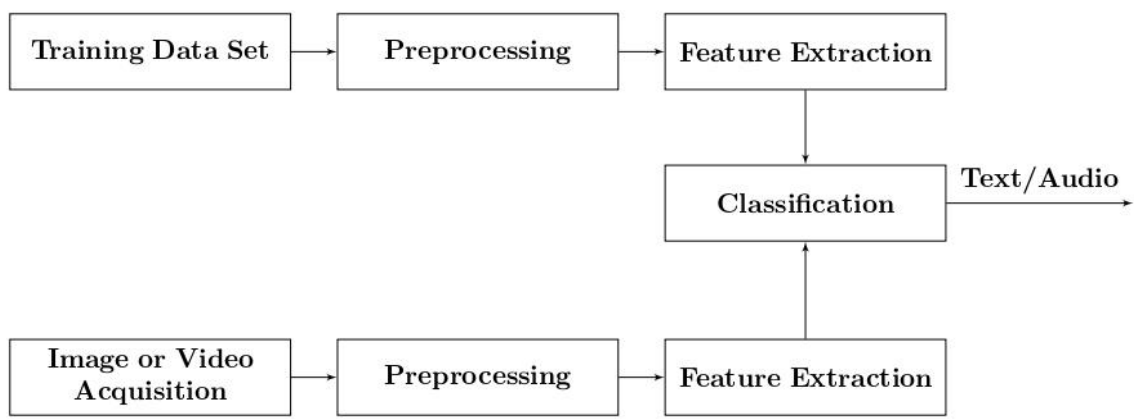

**Figure 1-2: Sign language recognition system** [3].

In [11], the system was created to recognize the Arabic sign words. The system consists of three stages, which are Pre-processing, Feature extraction, and classification. Moreover, the training and testing evaluation methods depend on the database of 23 signs that performed three signers. Each character is repeated 50 times by the singers, and the training set consists of 70%. The testing set consists of 30%. The model was evaluated on three different frequency domains (viz. Fourier, Hartley, and Log-Gabor transforms) for the feature extraction stage and assessed on three classifiers: KNN, SVM, and MLP. The results showed that the best Arabic sign language recognition system is Hartley transform using SVM classifier based on accuracy, 98.8%. Moreover, when sign images were segmented into 3*3 segments, the accuracy raised to 99%.

In this paper [12], it is focused on feature extraction. The feature extraction techniques are utilized for training the classifier. Sign language usually is dynamic where the upper part of the body, head, shoulder, and hands have a movement while other parts are static. The feature must be utilized by collecting high-contrast locations such as object edges and corners.

In [13], the authors designed a system to translate the ArSL alphabet gestures into text. The used dataset is captured from different smartphones by 30 volunteers. Each volunteer worked on a subset that has 30 images, so the dataset consists of 900 images. The authors used five descriptors to recognize. When using the Histograms of Oriented Gradients (HOG) descriptor, the proposed ArSL system accuracy is 63.56%. The accuracy of Edge Histogram Descriptor (EHD) is 42%. The accuracy of Discrete Wavelet Texture Descriptor (DWT) is 8%. The accuracy of the Local Binary Pattern (LBP) descriptor is 9.78%. The worst accuracy result is obtained using the 5 Gray-Level Co-occurrence Matrix (GLCM) descriptor; the proposed ArSL system accuracy is 2.89%.

The authors in [14] present a system that translates isolated Arabic word signals to text with automatic visual SLRs. The system consists of four stages to obtain the results: hand segmentation using the dynamic skin detector that depends on the face's colour, tracking using segmented skin points used to recognize and track the hands with the head's help, extracting the geo-metrical features of the spatial field. Finally, classification is carried out using Euclidean distance. As a result, the authors achieved a discrimination rate of 97% using a training set of 300 videos and a test set of 150 videos bearing in mind that 83% of words had different occlusions. These videos only contain 30 isolated words used in the daily life of hard of hearing children.

The authors in [15] present a new benchmark dataset publicly accessed along with the Sign Language Recognition algorithm. The SLR algorithm consists of three phases, which are hand segmentation, hand shape sequence, and body motion description, and sign classification. Also, the sign classification phase uses canonical correlation analysis and random forest classifiers. However, the dataset used for the algorithm was 150 different signs collected from 21 signers using the Kinect v2 sensor. The total sample is 7500 samples (150 signs * 5 signer groups * 10 samples per sign per group). Finally, the algorithm achieved a state-of-the-art solution when rated on the public data sets. Also, the achieved recognition accuracy is 55.57% evaluated on 150 ArSL signs.

The paper [16] starts sorting the sign language into two components; manual and non-manual signs. The manual signs include hand position, orientation, shape, and trajectory. The non-manual signs represent body motion and facial expressions. Convolutional Neural Network (CNN) is a deep learning class employed in image classification; it makes the network quick to learn and find the complex pattern simplicity. CNN still uses the Backpropagation and its derivatives training methods to learn from data. The author used a dataset of images containing 2030 images of numbers (from 0 to 10) and 5839 images of 28 letters of Arab sign language, i.e., 7869 RGB colour images with 256×256 pixels. These images are taken from different signers and different luminosity intensities.

The authors in [17] present an Arabic Sign Recognition system to overcome finger occlusions and missing data. The system uses two Leap Motion controllers for data acquisition since they detect hands and fingers moving. After that, data is put together using the Dempster-Shafer (DS) theory of evidence. A set of geometric features from both LMCs is chosen to feed them for the classification stage. Finally, the Bayesian approach with a Gaussian Mixture Model (GMM) and a simple Linear Discriminant Analysis (LDA) approach, used for classification. There are 2000 samples collected from two native adult signers by repeated 100 isolated Arabic dynamic signs ten times for each singer. Then, 70% of the dataset was used for training, and 30% used for testing. The submitted system is considered a state-of-the-art-glove-based system and single-sensor, and it achieved about 92% recognition accuracy.

The authors in [4] proposed a real-time ArSL alphabets recognition system. It consists of four convolution layers, four max-pooling layers, and five dropout layers. However, 54,049 images are used as a dataset for this system, consisting of 32 alphabets obtained from more than 40 participants. It is divided into 64% for training, 16% for validation, and 20% for testing. Finally, the achieved recognition accuracy was 97.6%.

The paper [18] shows a novel framework used to recognize isolated Arabic Sign Language words for signer-independent. This framework depends on three stages to classify input videos. These three stages are the DeepLabv3+ model used for hand semantic segmentation. The single-layer convolutional self-organizing map is used to extract hand shape features representation. A deep recurrent neural network is used to recognize the sequence of extracted feature vectors. The dataset comprises 150 repetitions for each of the 23 words they used, taken from 3 signers. In conclusion, the framework model achieves state-of-the-art performance with an average accuracy of 89.5%.

In [19], the authors exhibit sign language differently. Most researchers try to obtain the text rather than the semantic. To recognize the word with its semantic, they combined CNN with a semantic layer, and it maps the word to the meaning. A mobile camera picks the dataset in a different surrounding. The model achieved good recognition accuracy of 88.87%.

In [20], M. M. Kamruzzaman creates a model to convert Arabic Sign Language images to letters by CNN and then convert generated Arabic letters to Arabic Speech by Google Text to Speech. The CNN model has 2 Convolution layers, and the first layer has 32 Kernels, and the second has 64 kernels. The model also trained for 100 epochs on 100 images for every 31 letters and tested 25 images for each letter. It got an accuracy of about 90% for the testing set.

In [21], the authors focused on the recognition of letters. The collected data was 900 coloured images have been used to represent the 30 different hand gestures and have been used as a training set; another 900 images have been taken and used as a test set. They developed the recognition system and calculated its performance using Feed-Forward Neural Networks and various Recurrent Neural Networks (RNN) types. The performance that they got is concluded in Table 1-1.

**Table 1-1: Proposed system with different classifiers** [21].

| Classifier | Accuracy |
|---|---|
| Feed-Forward Neural Network | 79.33% |
| Elman neural network | 89.66% |
| Jordan neural network | 84.56% |
| Fully recurrent neural network | 95.11% |

The authors in [22] proposed the first ArSL recognition system that converts ArSL to Arabic sentences. The machine translation system is Rule-based, and it has three stages; the input Arabic Sign language word is processed for Morphological analysis then Syntactic analysis. Finally, transfer to Arabic sentences. However, the system used a corpus that has sentences that are used in health centres. It has 600 sentences that consist of 3327 sign words with 593 unique sign words. The proposed dataset is divided into training, validation, and testing datasets, with a percentage of 70%, 15%, and 15%, respectively. The results of the system are calculated Manually and automatically. However, the manual evaluation shows that 80% of the sentences are accurately translated, and 2 ArSL experts do the evaluation. Also, it is evaluated automatically by BLEU and TER metrics and gets 0.39 and 0.45 repetitively.

## 1.4 Aims and Objectives

Facilitation of deaf people's lives and making their communication more straightforward is what we aim to achieve. The objective is to construct a simple link between deaf people and others by creating an accurate automated model using deep learning to interpret sign language alphabets to text. We will study the previous results that the others obtain, enhance the model's performance, and make a prototype to test the model. In the future, we will do continuous feedback to diagnose any error and fix it. We will expand our work to include the words and deal with full sentences.

## 1.5 Report Organization

In the rest of this work is organized as follows: Chapter 2 (Design Considerations), Chapter 3 (Model architecture, Training and testing), Chapter 4 (Design Testing and Results), Chapter 5 (Conclusion And Future Work).

# Chapter 2 DESIGN CONSIDERATIONS

This chapter explains the software, frameworks, techniques, and alternatives that can also be needed in the project. Also, we will show design constraints and standards.

## 2.1 Design Options

Image recognition requires complex calculations to accomplish it using the computer. So, we need powerful techniques and appropriate software to achieve it. In general, computer vision can do this smoothly, but not all computer vision techniques are suitable for image classification. Also, many software and frameworks can be used in computer vision.

### 2.1.1 Computer Vision Techniques

#### 1. K-Nearest-Neighbor Classification (KNN)

In [23]. K-Nearest Neighbour is considered supervised learning in which the features and labels are given in the model. This technique depends on the closest distance between the point and the predicted labels to classify the object. An unlabelled query point is assigned the label that has the K-Nearest Neighbour. The classification process is calculated from most of its K Nearest Neighbours. To classify the images, each image is converted to a fixed vector, then the distance can be measured by any function; Euclidean distance is the most common function:

$$d(x,y) = \|x - y\| = \sqrt{(x-y) \cdot (x-y)} = \sqrt{\sum_{i=1}^{m}(x_i - y_i)^2} \qquad (2.1)$$

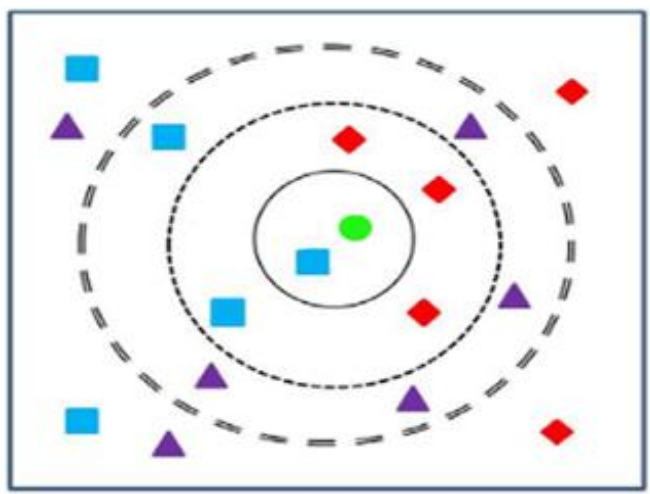

**Figure 2-1: KNN Classification** [23].

## 2. Linear Classifiers

"Linear classifiers classify data into labels based on a linear combination of input features. Therefore, these classifiers separate data using a line or plane or a hyperplane. They are suitable to classify the linear separable data." [24]

### 2.1 Logistic Regression (Binary Classification):

A statistical model can be used to evaluate (guess) the probability of an event depends on input data.

**For example**, we have two classes, e.g., "dog" or "not dog" and those can be represented by 0 and 1.

It can be mathematically represented as follows: $\hat{y} = \sigma(z)$

$$\sigma(z) = \frac{1}{1 + e^{-z}}$$

(2.2)

is the logistic function and

$$z = w^T x + b$$

(2.3)

And these parameters are as follows:

$\omega$ : weight

$b$ : bias

$x$ : flattened feature input vector

The model takes x as an input, and the probability of the outputs $\hat{y} = \rho(y = 1|x)$

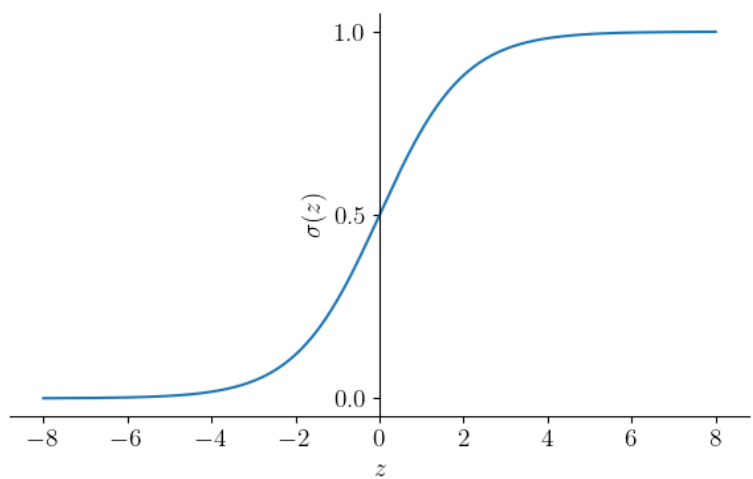

**Figure 2-2: Sigmoid Function (Logistic Function)**

For the images x, the feature vector can be just the pixels' values in RGB channels, and it can represent by a vector with one dimension. It can be resulted by flattening those three dimensions, and the resulted size is $n_x = n_{height} \times n_{width} \times 3$.

The goal of this algorithm is to classify the images correctly, and this can do by training the model on training samples that will change the values of $w$ and $b$. The optimal values of these parameters can be justified when $\hat{y}^{(i)}$ most closely predicts $y^{(i)}$. Where:

$\hat{y}^{(i)}$ : predicted class value.

$y^{(i)}$ : correct class value.

In practice, this model usually calculates the loss function:

$$L\left(\hat{y}^{(i)}, y^{(i)}\right) = -[y^{(i)} \log(1 - \hat{y}^{(i)}) + (1 - y^{(i)}) \log(1 - \hat{y}^{(i)})] \tag{2.4}$$

For each training example and minimizing the cost function,

$$J(w, b) = \frac{1}{m} \sum_{i=1}^{m} L\left(\hat{y}^{(i)}, y^{(i)}\right) \tag{2.5}$$

Overall m training examples.

$$\frac{\partial L}{\partial w_j} = \left(\hat{y}^{(i)} - y^{(i)}\right) x^{(i)}_j \text{ and } \frac{\partial L}{\partial w} = \left(\hat{y}^{(i)} - y^{(i)}\right) \tag{2.6}$$

Where $j = 1, 2, \dots, n_x$ labels the components of the feature vector.

To get the optimal value of $w$ and $b$ ; $J$ should be minimized. It can be minimized numerically after choosing initial values by changing them according to descent along the steepest gradient.

$$w_j \rightarrow w_j - \alpha \frac{\partial j}{\partial w_j} = w_j - \frac{\alpha}{m} \sum_{i=1}^{m} \frac{\partial L}{\partial w_j} \tag{2.7}$$

$\alpha$ is the learning rate (step size), which affects how large each step is taken in the direction of greatest decrease in $J$. Choosing a good value for α is a subtle art (where the too-large value will affect the training to be fast and the training may not converge steadfastly and too small value so the training will be slow).

### 2.2 Softmax and SVM classifiers:

The linear classifier uses the below equation to learn the features of images and stores them in W, b:

$$f(x_i, W, b) = W\,x_i + b \qquad (2.8)$$

$W$ : Weights.

$b$ : bias term.

$x_i$ : input image.

$f(x_i, W, b)$ : Score function.

W, b (module parameters) will be changed depending on the training dataset. The output module will classify the image depending on its features (pixel value), and space will be divided by linear functions [25].

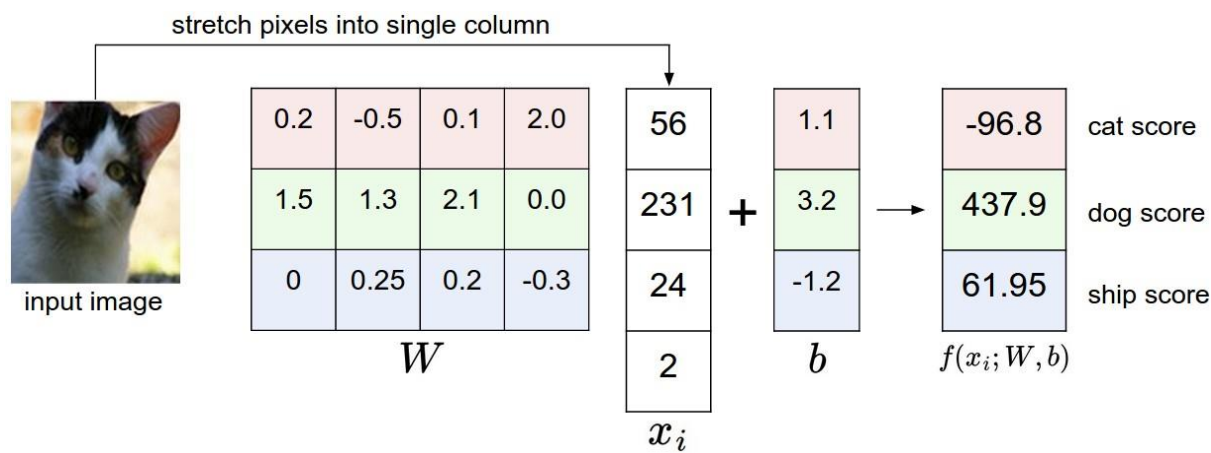

**Figure 2-3: Mapping a Cat Image to Class Scores** [26].

The above image is added as an example to clarify the idea of the linear classifier. For ease of visualization, the image is assumed to have 4 pixels only. And $W$ is considered as a matrix with a size of $3 \times 4$, where 3 is the class number, and 4 is the flattened input size to imply the matrix multiplication between $W$ and $x_i$. So, by doing the matrix multiplication of $W$ and $x_i$ then adding the bias $b$ so the results will be the scores for each class. The $W$ in the image is bad, and the scores at the end claim that the image is a dog not a cat. However, the $W$ will improve by train the model, and it may get better results.

There are many loss functions that can be used. For example, the linear classifier model usually uses a loss called the **Multiclass Support Vector Machine** (SVM) loss. So, the Multiclass loss can be shown as below:

$$L_i = \sum_{j \neq y_i} \max(0, S_j - S_{y_i} + \Delta)$$

(2.9)

Where:

$S_j$ : is the score for the j[th] class.

$S_{y_i}$ : is the score for the i[th] class.

$\Delta$ : is the fixed margin.

Another commonly used classifier is Softmax, which used **cross-entropy loss**. The function mapping is still used $f(x_i; w) = wx_i$ . And the **cross-entropy loss has the following form:**

$$L_i = -\log\left(\frac{e^{f_{y_i}}}{\Sigma_j e^{f_j}}\right)$$

(2.10)

Where $f_{y_i}$: is the class score for the i[th] element.

Where $f_j$: is the class score for the j[th] element.

And $\frac{e^{z_j}}{\Sigma_j e^{z_k}}$ It is called Softmax Function, which has an input vector score (in z) squishes it to a vector of values between zero and one (Probability), which sum to one.

The below images show the difference between SVM and Softmax classifier for the same input. Both have the same mapping function, which resulted from the matrix multiplication. But there is a difference in the interpretation of the score function. SVM interprets the class scores, and it encourages the correct class to be the higher one by a margin than the other classes.

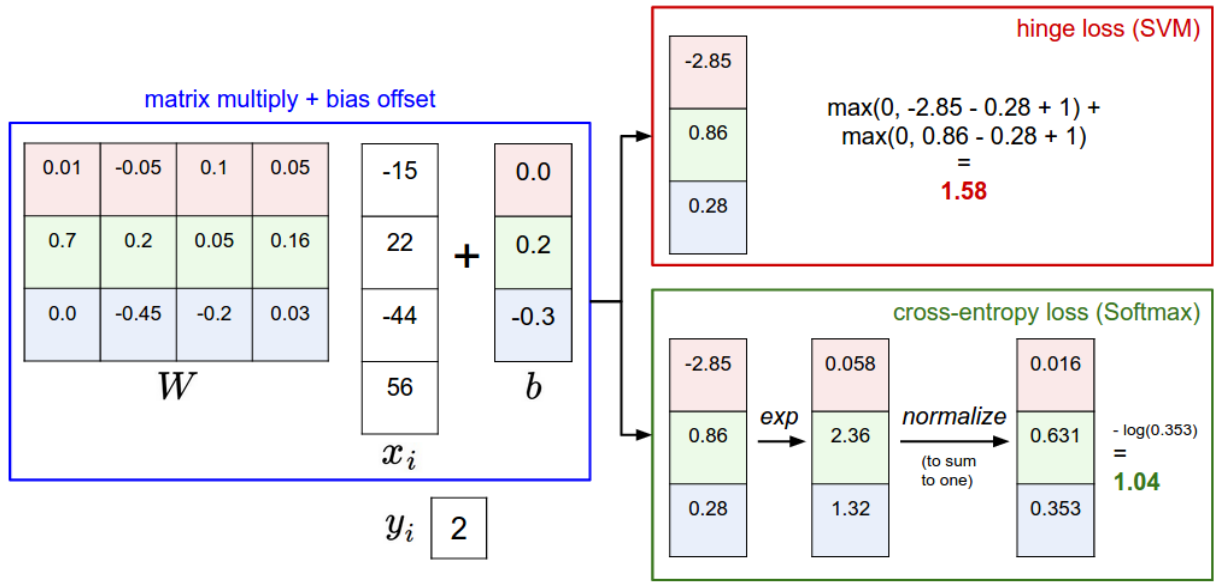

**Figure 2-4: Example for Softmax and SVM** [26].

## 3. Artificial Neural Networks

ANNs are simulated based on the brain's architecture. They consist of connected elements known as an artificial neuron; it has one or multiple inputs and one output with either zero or one. Each neuron is associated with a weight, and if their sum is more than the threshold, the neuron will activate. The following equation will explain the mathematical representation:

$$x = \begin{cases} 1, & \sum_i w_i x_i - b > 0 \\ 0, & \sum_i w_i x_i - b \leq 0 \end{cases} \tag{2.11}$$

If we want to make the output smoother between zero and one, we will use the sigmoid function as the following:

$$\sigma = \frac{1}{1 + e^{\sum_i w_i x_i - b}} \tag{2.12}$$

Figure 2-5 shows a simple neuron with three inputs associated with its weights and the bias and then applying the activation function to show the result.

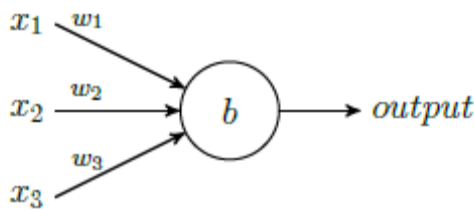

**Figure 2-5: Simple Neuron** [27].

ANNs consist of layers as an input layer, one hidden layer or more, and an output layer. Each layer consists of neurons that compute the weighted sum of their inputs then specify the output using some of the activation functions like; sigmoid, ReLU, and SeLU. All the neurons in a layer are considered an input to the followed layer. ANNs can recognize the output by modifying the weights and biases each one epoch until minimizing the errors. We need to classify the result of each neuron in the output layer to the predicted class. So, Softmax function can be used at the output layer. Figure 2-6 shows the architecture of ANN that includes ReLU and Softmax.

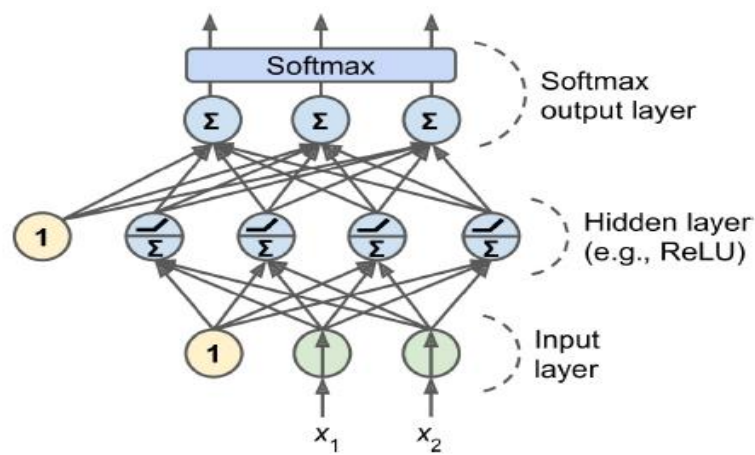

**Figure 2-6: ANN Architecture, Including ReLU and Softmax** [28].

### 4. Convolutional Neural Networks

CNN is the most dominant technique in deep learning that can use in computer vision tasks. CNN is a mathematical model that consists of three types of layers as follows:

#### a) Convolutional Layer

It is an essential component of the CNN architecture used for feature extraction. Neurons in one layer are connected to other neurons in their receptive field. The array that combines the neurons is called the kernel, and it is typically formed as $3 \times 3$, but maybe choose $5 \times 5$ or $7 \times 7$. This architecture allows the low-level features to concentrate on one layer, then assemble them into higher-level features in the next layer.

The operation above does not guarantee each kernel's centre to overlap the input layer's outermost element. Padding, precisely Zero Padding, is a solution to avoid adding zeros around the inputs that can overlap the outer element of the input layer.

Stride is "the distance between two consecutive Kernels.", the standard option of a stride is one. Figure 2-7 shows an example of a convolutional operation with a kernel size $3 \times 3$, Zero Padding, and a stride of one.

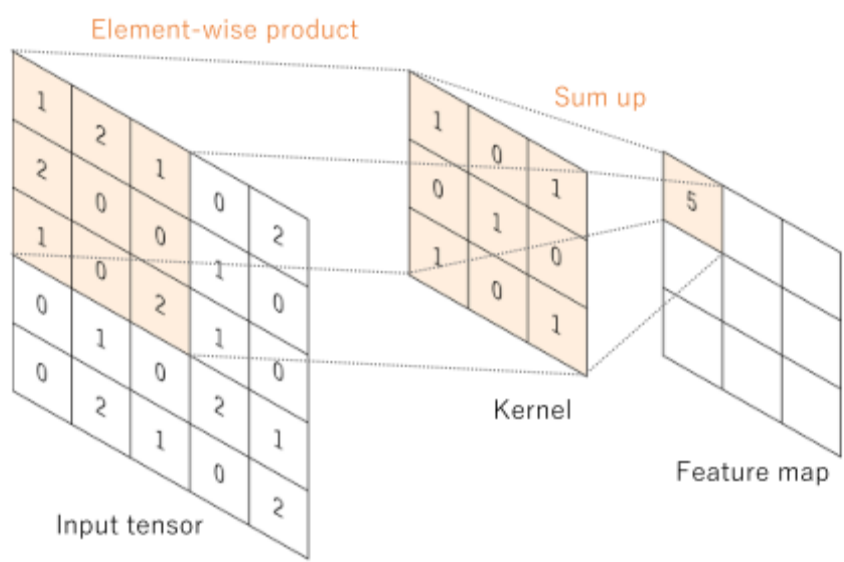

**Figure 2-7: An Example of A Convolutional Operation** [29].

#### b) Pooling Layer

The goal of this layer is to shrink the inputs to decrease the computations. Max pooling and Mean pooling are common examples of pooling layer. Max pooling is the most popular form, which takes the maximum value in the higher-level feature layer. Mean pooling takes the average of all the elements in the higher-level feature layer.

**c) Fully Connected Layer**

This layer transforms the last convolutional layer into a one-dimensional array and connects to one or more dense layers. A non-linear activation function follows the final fully connected layer to classify the inputs according to the output probabilities.

**5. Transfer Learning**

Many computer vision cases have small datasets, so the training of the model will be invalid. The popular approach to deal with this case is to use the transfer function. Transfer learning is a network that comprises extensive data, and it was trained to absorb generally feature extraction of the image classification task. Convnet, VGG16, ResNet, Inception, and Xception are examples of architectures trained on ImageNet (1.4 million labelled images with 1,000 different classes). It is preferred to choose the understood architecture for you, and no need-to-know new ideas.

## 2.1.2 Software

**1. MATLAB**

It is a programming platform that offers toolboxes to help engineers and scientists in academia and industry to perform the solutions for various aspects of problems. The essence of MATLAB is a matrix-based language that allows for progressing the calculations smoothly. MATLAB can deal with data by analysing and visualizing it, improving existing algorithms to coincide with your requirements, and creating models and apps from scratch [30]. MATLAB includes Many applications and capabilities that can perform several functions as follows:

**1- Applications** [31]**:**
    **a. Image Processing and Computer Vision:** Processing of images and videos using several techniques to build any visual model.
    **b. Data Science:** Use machine learning to predict and label the data.
    **c. Deep Learning:** Apply deep neural networks and prepare the related data.
    **d. Signal Processing:** Convert the signal and prepare it to analyse.

**2- Capabilities** [31]**:**
    **a. Algorithm Development:** Improve or build algorithms for several tasks.
    **b. Cloud Computing:** Run public clouds like; AWS and AZURE on MathWorks cloud.
    **c. Data Acquisition:** Gain the data from an external source like a camera.
    **d. GPU Computing:** Offer using NVIDIA CUDA to accelerate the training.
    **e. Parallel Computing:** Use CPUs, GPUs, and TPUs simultaneously in large systems.
    **f. Real-Time Simulation and Testing:** Apply the hardware systems in real-time.

MATLAB is a useful software for Machine Learning because of its simplicity of use and offering toolboxes that support machine learning algorithms. The toolboxes like; image processing and computer vision, data science, and deep learning include all the tools to train and test the models. MATLAB offers parallel computing to operate CPUs, GPUs, TPUs, and clouding to achieve high performance [32].

## 2. LabVIEW

LabVIEW is software designed for engineering problems that require the acquisition of the data, testing it, measuring it, and controlling it. The most feature of LabView is its ability to create mutuality environment between the hardware and data insight. LabVIEW provides the user with a graphical programming approach, toolkits, and modules that help the user visualize any application like working in a real lab, including hardware configuration, instrumentation to measure the data, and error debugging. This integration between hardware and software can simplify building a complex diagram and applying it on hardware, improving the data algorithms, and designing special user interfaces [33].

LabVIEW contains Analytics and Machine Learning Toolkit that combines predictive analytics and machine learning. The toolkit is prepared to deal with large data and do some processes like, classification, clustering, and anomaly detection. And it has good advantages which are to monitor the conditions and maintain the predictive [34]. Figure 2-8 shows some of the processes can LabView applying on data to get some results.

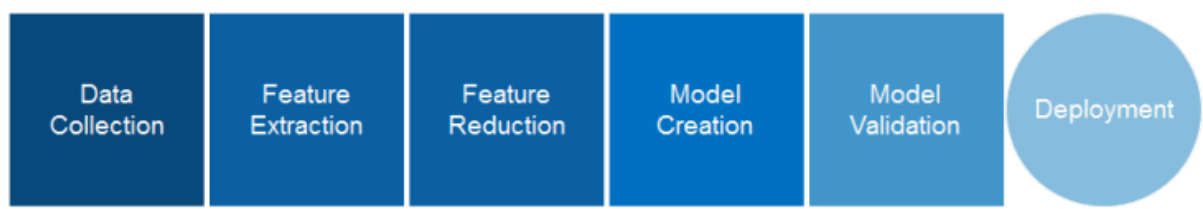

**Figure 2-8: Machine Learning Processes Which LabVIEW Can Provide** [35].

There are some explanations on the processes:

a. **Data Collection:** DAQs (Data Acquisitions) allows picking up the required data.

b. **Feature Extraction:** Some tools like; Vision Development Module, NI Sound and Vibration Measurement Suite can extract the features from the data based on your domain knowledge.

c. **Feature Reduction:** Use some techniques to simplify the data and reduce its dimension to prepare it for training.

d. **Model Creation:** Give the flexibility to build and train the models.

e. **Model Validation:** Use evaluation metrics to check the validation of the models.

f. **Deployment:** Use deployment data to predict new data.

**3.  Julia**

"It is a high-level, high-performance dynamic language, focusing on numerical computing and general programming." Traditional computing languages were either fast or productive, but not both. Julia achieves fast and productivity [36].

Julia contains packages supporting computer vision tasks and includes open-source libraries like Open CV and Tesseract to find optimum computer vision tasks. Julia can deal with simple images using Julia Images to advanced Images using Julia's APIs [37].

**4.  Python**

"Python is an interpreted, object-oriented, high-level programming language with dynamic semantics," Python is easy to learn because it supports code readability and therefore reduces the bugs. Python consists of dynamic typing and dynamic biding that make the program shorter and faster. Python supports packages with a wide range of functionality like; data analytics, databases, graphical user interface, image processing, and scientific computing, which allows the code to be reused and decreased the effort required to build the code from scratch [38].

Python is considered the most common programming language for machine learning and data science because it allows forgetting the complex parts of programming by putting the concepts directly into the goal. Python provides us with many libraries and frameworks that offer loading data, prepare data, label data, visualize data, and apply the different algorithms to train and test the models [39].

### 2.1.3 Python Frameworks

A framework is an interface that makes machine learning models simpler and speeds up the processing of models. Frameworks allow connecting the data with the models as APIs and observe your model and its performance. The famous frameworks that are used in Python:

1. **Scikit-Learn**

It is an open-source machine learning framework that implements various model fitting functions, data extraction, and many other advantages. It is straightforward to use so. It is considered an entry point to enter the machine learning field [40].

2. **TensorFlow**

It is an advanced open-source framework that can achieve complex computations. It allows us to build huge flexibility models because it has a rich library that contains many functions and prepared models. TensorFlow offers cloudy hundreds of GPU servers [41].

3. **Keras**

It is a high-level Deep Learning API (Application Programming Interface) that can simplify building the model and training it. Reducing cognitive load is considered one of TensorFlow's most feature, which can load data efficiently [42].

4. **PyTorch**

It is an advanced open-source framework that has tools to improve computer vision and reinforcement learning fields. It provides cloud platforms and the ability to use GPUs to accelerate the models [43].

### 2.1.4 Hardware

Deep learning algorithms like; computer vision or automatic speech recognition require computational power because the model becomes deeper and has big data to analyse. Many hardware units can deal with big data and reduce the training time as follows:

1. **Central Processing Unit (CPU)**

It is an integrated circuit that performs machine instructions using arithmetic, logic, controlling, and input/output operations stated by the commands. CPU includes an Arithmetic Logic Unit (ALU), Central Unit (CU), and Memory Unit (MU). ALU is used to execute arithmetic and logical operations. CU uses the data bus and control bus to organize the control signals. MU includes the various aspects of memory such as Random-Access Memory (RAM), Read-Only Memory (ROM), and CACHE. CPU performs the operations, where registers are loading the values and storing them, CACHE memory retrieving the values, CU organizes the requests and controls the priorities steps to process the input according to the ALU requests. Figure 2-9 shows the principal components of the CPU [44].

Most modern CPUs are embedded in IC chip that includes the CPU, memory, and peripherals. Modern CPUs have multi-cores, where each core can run several threads. Most Machine Learning algorithms are based on matrix multiplications and additions so, CPUs cannot quickly achieve this arithmetic calculation; for example, training a deep network with a single chip can continue for days or weeks. The Frameworks can operate CPU and GPU parallel; the heavy computations on the GPU, and data processing on the CPU [28].

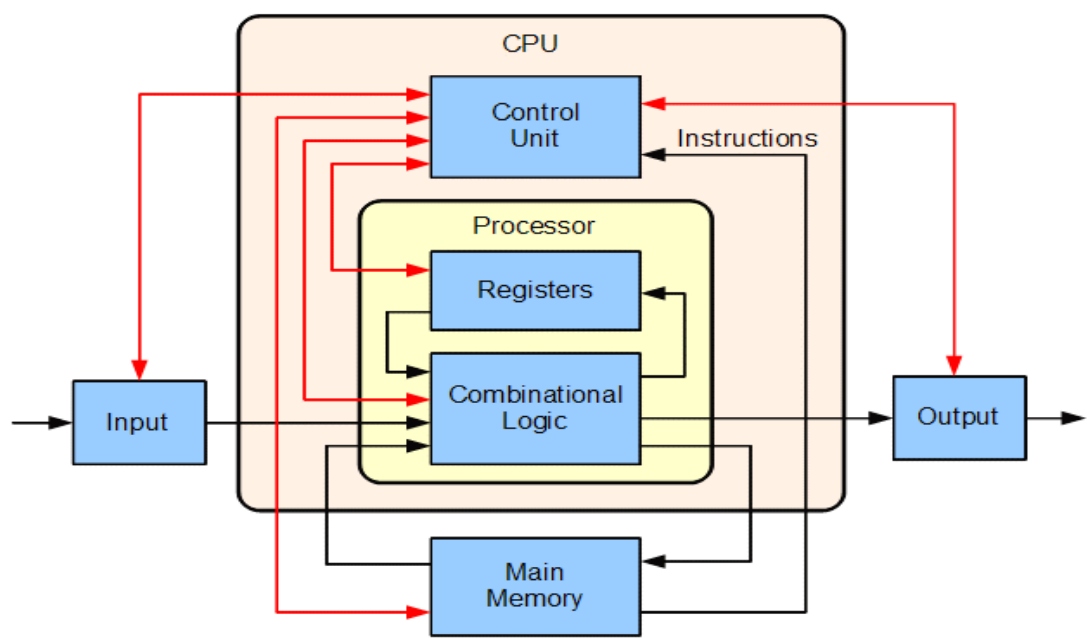

**Figure 2-9: CPU Operations** [44].

## 2. Graphical Processing Unit (GPU)

GPUs become an essential integral part of computing's systems because they become more complex and need to be faster with high-performance in various aspects like, gaming and Machine Learning applications. The GPUs are the best choice for large computations applications [45]. GPU is a high-computational performance processor for graphical processing. GPU was designed for parallel processing and high memory bandwidth to accomplish high computational power and increased productivity. The GPU's architecture essential component is the Streaming Multiprocessor (SM), also called CUDA-Cores by NVidia. SMs contain many ALUs, and each SM can operate one warp (a pack of 32 threads) simultaneously [27].

3. **Tensor Processing Unit (TPU)**

"It is custom-developed application-specific integrated circuits (ASICs) used to accelerate machine learning algorithms." Cloud TPUs allow us to train the models on TensorFlow with high performance and less time. Machine learning's essence is the mathematical computations that minimize the error between inputs and predicted outputs, so cloud TPUs accelerate calculations' performance. It is advised to use cloud TPUs in these cases; the models are constructed from matrix computations, the models that require weeks or months for training, and the large models that contain more and more layers with huge batch size [46].

## 2.2 Design Constraints and Standards

Constraints are restrictions that prevent something from being the best. They can be problems that arise or issues that come up. Some constraints must be considered in our project as follows:

1. **Availability of Data:**

   The AI, Machine Learning and Deep Learning models are hungry for data. Especially, Deep Learning models need more than 1000 images for each class. And those images should agree with the real world with many backgrounds, noise, illuminations, and the direction of the image. And there are few resources for Arabic sign language images. So, we need large data with various aspects to get a good model with high accuracy and cover all the possibility's images. Also, some resources are not available for everyone.

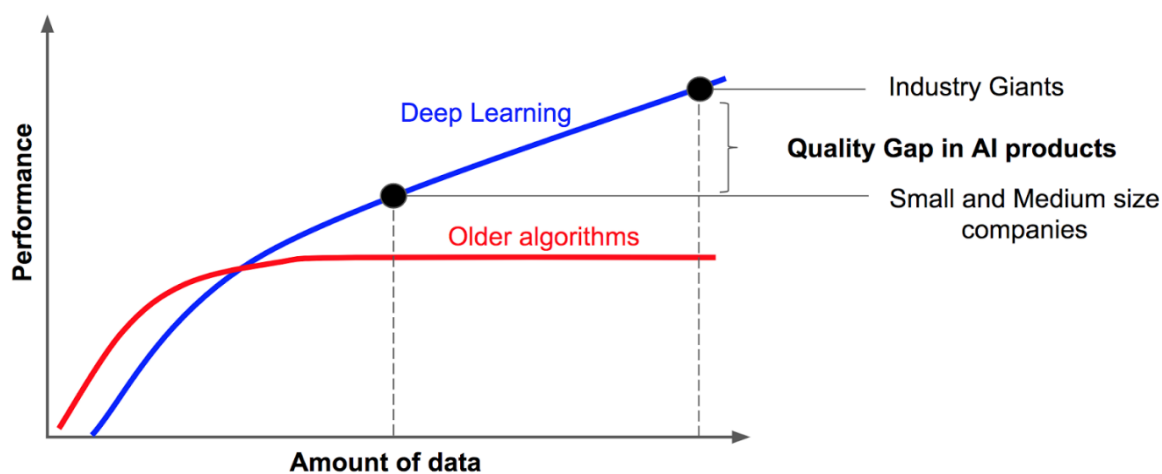

**Figure 2-10: Performance Vs Amount of Data** [47].

2. **Computational Resources:**

Training of large-scale data usually needs GPUs or TPUs to accomplish the computations and memory usage. Also, it may need many GPUs to be used at the same time. Therefore, this will affect the time required to get results and test the model for many cases to check its performance. However, these resources are expensive to afford for model training.

3. **Response time (Inference Time):**

Real-time systems required short inference time to deploy the model for real-life usage with minimum computation resources.

4. **Hyperparameter Choosing:**

These are considered critical for every model. And choosing these parameters is subtle art rather than standard choosing. However, you can use similar research and models hyperparameter as a guide for your research.

5. **Knowledge and Experience in ArSL:**

This project is a multi-disciplinary project that combines computer vision with ArSL. So, it needs an expert in ArSL to take care of the Arabic sign language part of the project.

# Chapter 3 MODEL ARCHITECTURE, TRAINING AND TESTING

This project aims to build a model that can convert the alphabetical image in ArSL to the corresponding written letter in the Arabic language by applying some algorithms that can extract features and differences in the image. Choosing the right algorithm depends on the nature of data, complexity, and required resolution of the images. This model uses the PyTorch framework to complete this task by testing several models and comparing final results based on the ArASL2018 dataset [48].

Usually, it would be best to try many approaches before getting the best one for a new problem. Even experienced machine learning researchers need a lot of ideas before getting the desired results. An experienced AI engineer suggest following a cycle to get satisfactory results, as follows [49]:

1. Create an idea on how to build the model.
2. Convert the idea to code that can be implemented.
3. Evaluate the idea by an experiment.
4. Terminate or go back from the beginning and generates more ideas, then keep this continues iterating.

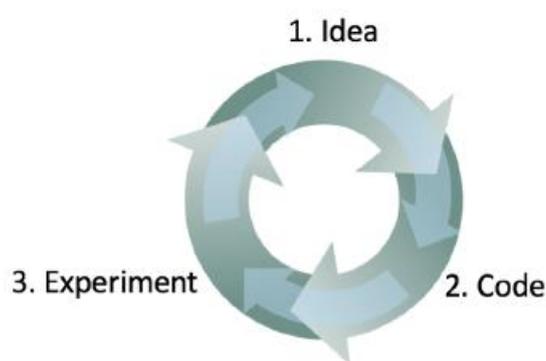

**Figure 3-1: Machine Learning System Cycle** [49].

Before constructing any code, we should do a general flowchart or pseudocode to plan the model's steps to know the path on it to achieve the task. It helps to be systematically through programming the code. Figure 3-2 shows the general flowchart for the project:

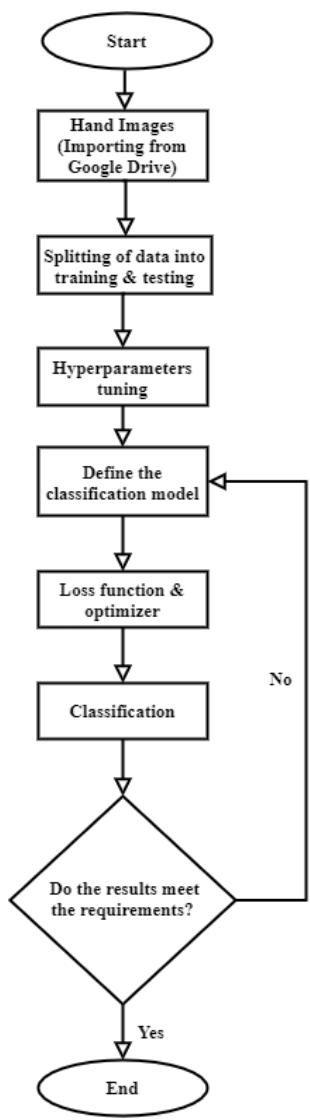

**Figure 3-2: Flowchart of Recognition Model**

The explanation of the flowchart will be in the following steps:

- **Importing of Data:** The model will be trained using Google Colab, and we need to feed the data into the model. One approach to importing data is uploading locally from PC, but this way is not proper because we will need to upload the data again every time opening Colab. Another approach is uploading data to Google Drive once, and we just recall the data when we need it.

- **Splitting Data:** We need to split data perfectly into training, validation, and testing datasets. These datasets assure to generalize the data and examine the model performance. Scikit-learn supplies functions that can split the data perfectly with various features like; random state parameter that generates random datasets and gives the same dataset if you select the same seed. Another feature is splitting the data into subsets with the same indices. The most important feature is stratified sampling which means splitting the data into stratified datasets, which mean that datasets have the same proportion of input dataset.

- **Hyperparameters Tuning:** The models in deep learning have several parameters like; the number of epochs, the number of batches, learning rate, and dropout. Parameter's manipulation can make the model better or worse, and selecting parameters depends on the experience or testing several trials until satisfying the results.

- **Defining the Model:** It means constructing the model's architecture, including the type of layers, number of layers.

- **Loss Function and Optimizer:** To assist the model, we need a loss function to ensure that the training is doing well. Also, we need an optimizer to update the weights inside the network.

- **Classification:** At this stage, the model is trained, validated, and tested. After that, we see the results if they satisfy our target or not.

## 3.1 Dataset Examination

### 3.1.1 The Nature of Images

In terms of colour, it can be in grayscale representation where it has 1 channel or in RGB representation where it has 3 channels. [50]

- **Grayscale:** It contains shades of grey, proportional to the pixels' luminance. The luminance index changes from 0 to 255, examples of these are black, which is 0, and white which is 255. Figure 3-3 (a) shows an image that is represented in grayscale.

- **RGB:** The colour of each pixel in the image can be represented as a combination of RGB space colour, and this allows the user to specify the colour intensity between 0 and 255 for each channel, and the combination of this mixture determines the final colour, examples of which are RGB (255, 0,0) is red, RGB (255,255,0) is yellow, and RGB (128,0,128) is purple. These pixels can be combined to form the image we see in real life. Figure 3-3 (b) shows an image that is represented in RGB colour space.

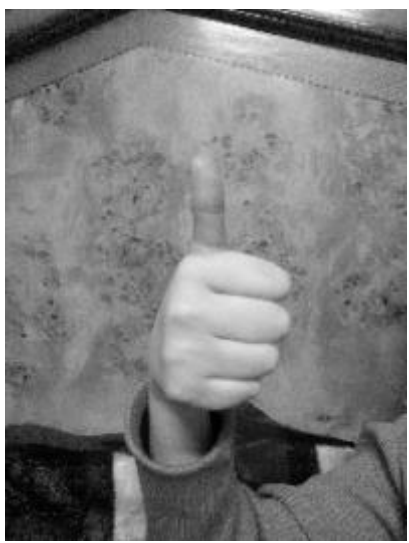    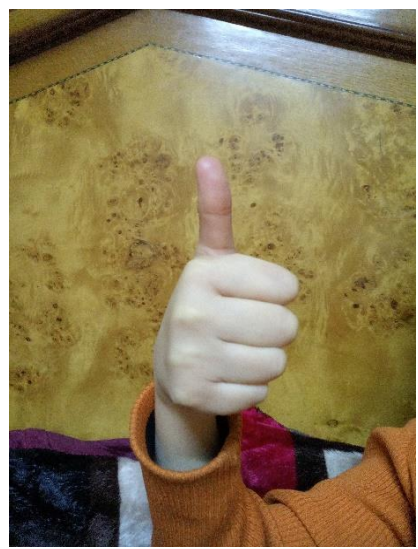

(a)                                                              (b)

**Figure 3-3: (a) Grayscale Colour Space, (b) RGB Colour Space**

Our project has 54,049 images in grayscale, so that the model will be trained and tested based on grayscale colour space. The computations in grayscale will be simpler than RGB because we deal with 2D arrays, making the model faster.

## 3.1.2 ArASL2018: Arabic Alphabets Sign Language Dataset

ArASL2018 is a dataset that contains 54,049 fully labelled images for 32 alphabets in Arabic sign language. However, these images are performed from 40 participants of different ages. These images are grayscale with $64 \times 64$ pixels JPG format that captured using a smart camera. [4]

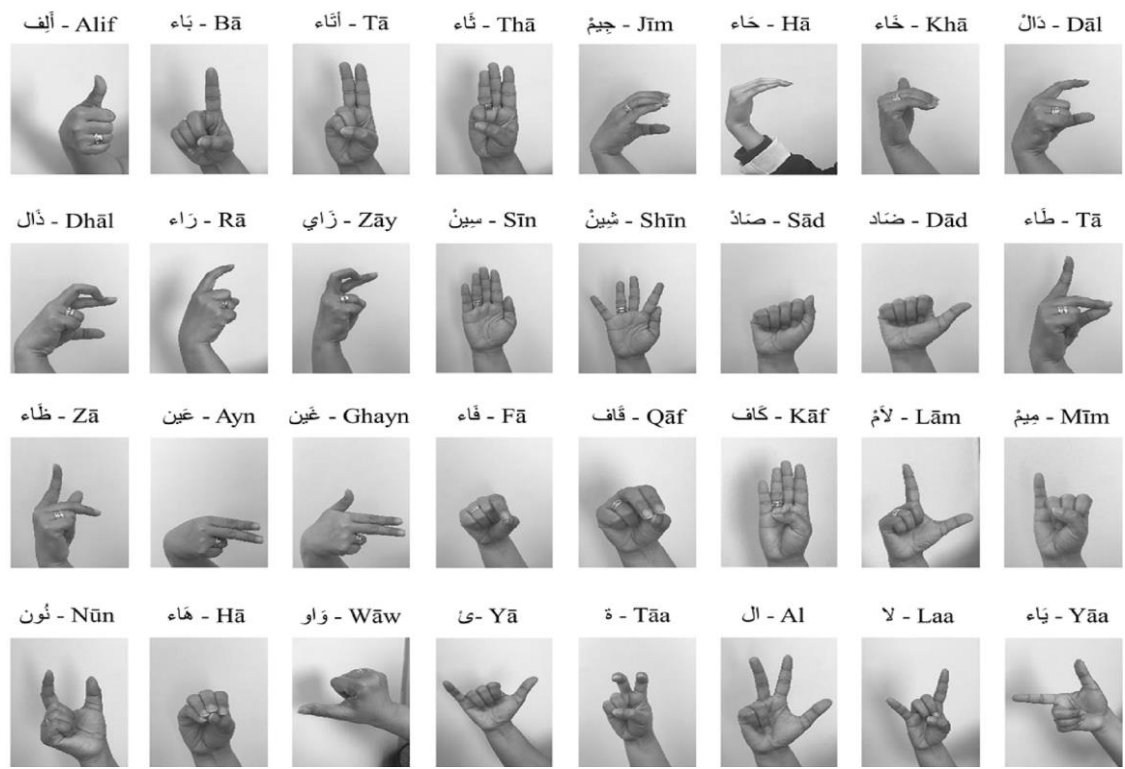

**Figure 3-4: Samples For ArASL2018 Dataset** [4].

| # | Letter name in English Script | Letter name in Arabic script | # of Images | # | Letter name in English Script | Letter name in Arabic script | # of images |
|---|---|---|---|---|---|---|---|
| 1 | Alif | (ألف)ا | 1672 | 17 | Zā | (ظاء)ظ | 1723 |
| 2 | Bā | (باء)ب | 1791 | 18 | Ayn | (عين)ع | 2114 |
| 3 | Tā | (تاء)ت | 1838 | 19 | Ghayn | (غين)غ | 1977 |
| 4 | Thā | (ثاء)ث | 1766 | 20 | Fā | (فاء)ف | 1955 |
| 5 | Jīm | (جيم)ج | 1552 | 21 | Qāf | (قاف)ق | 1705 |
| 6 | Hā | (حاء)ح | 1526 | 22 | Kāf | (كاف)ك | 1774 |
| 7 | Khā | (خاء)خ | 1607 | 23 | Lām | (لام)ل | 1832 |
| 8 | Dāl | (دال)د | 1634 | 24 | Mīm | (ميم)م | 1765 |
| 9 | Dhāl | (ذال)ذ | 1582 | 25 | Nūn | (نون)ن | 1819 |
| 10 | Rā | (راء)ر | 1659 | 26 | Hā | (هاء)ه | 1592 |
| 11 | Zāy | (زاي)ز | 1374 | 27 | Wāw | (واو)و | 1371 |
| 12 | Sīn | (سين)س | 1638 | 28 | Ya | (ياء)ئ | 1722 |
| 13 | Shīn | (شين)ش | 1507 | 29 | Tāa | (ة)ة | 1791 |
| 14 | Sād | (صاد)ص | 1895 | 30 | Al | (ال)ال | 1343 |
| 15 | Dād | (ضاد)ض | 1670 | 31 | Laa | (لا)لا | 1746 |
| 16 | Tā | (طاء)ط | 1816 | 32 | Yāa | (ياء)ياء | 1293 |

**Figure 3-5: A Brief Description of Arabic Sign Classes** [4].

- **Problems in The ArASL2018 Dataset:**
  - There are different sizes of images in the dataset, which are:
    - The number of images with size $64 \times 64$ are 53401.
    - The number of images with size $1024 \times 768$ are 10.
    - The number of images with size $256 \times 256$ are 638.
  - These images are collected from a video, which means they are similar in the surrounding conditions and do not represent the actual life datasets, and this does not guarantee the variation of data to explore the pattern of images.
  - Some of the images in the dataset need the motion to describe the sign entirely so, training of them will be useless without the movement.
- **Our Solution for The Problems:**
  - We resized all images to $224 \times 224$.
  - In this stage, we used the available images with greyscale, but we collected new dataset that has new features.

## 3.2 Data Splitting

To build a suitable model in machine learning, we need to split the dataset into three sets: training, validation, and testing. The model uses training dataset to tune the parameters, weights, and biases. Also, testing and validation datasets are the best way to check how well the model can deal with new data and generalize the new cases.

The validation and testing datasets usually are from the same distribution that the model with work with in the future, but this is not necessary for the training dataset.[49]

There are some explanations about the three sets briefly as follows:

- **Training Dataset:** Instance data that the system can predict patterns by finding the optimal parameters to obtain a good prediction on new data. The training process is like humans learning, where people can learn complicated techniques by analysis of repeated patterns.
- **Validation (Development) Dataset:**
  - It is used to get the right parameters through tuning the networks.
  - It is used to solve the overfitting issue; we take an instance of data as a validation dataset to compare models.
- **Testing Dataset:**
  - It measures the performance of the network but without making any tunning on the model's parameters.

A high percentage of data goes to training as 70%, and validation as 15%, and testing as 15%. These percentages can differ according to the dataset's size. Usually, the data cannot be generalized, or we need to enlarge the dataset so, we manipulate and transform the dataset by, flipping, cropping, and resizing. This process is called data augmentation. Figure 3-6 shows the processes on the dataset briefly.

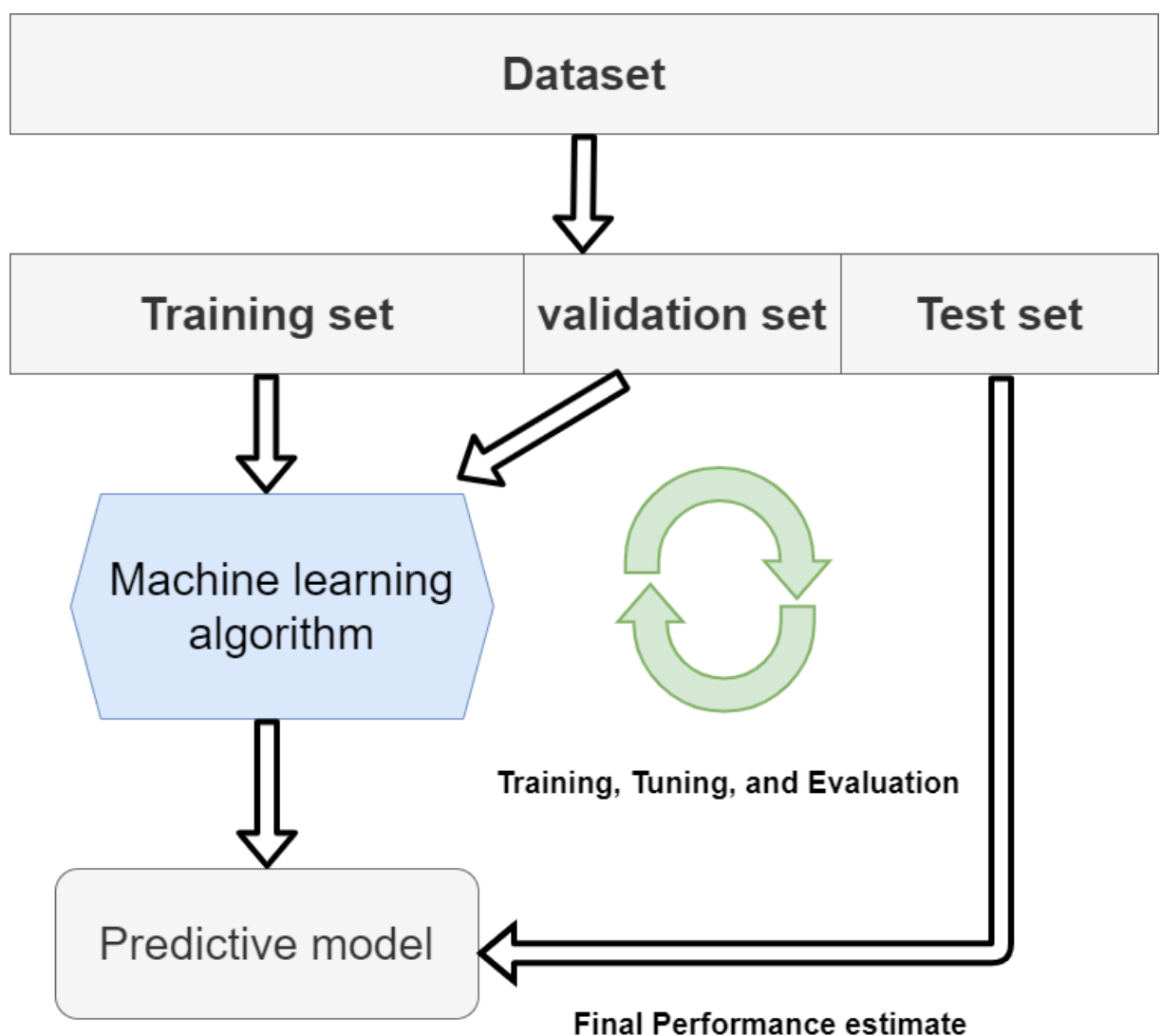

**Figure 3-6: General processes on the dataset**

The data are distributed to 32 classes, as shown in Figure 3-7, and we need to split it into three sets: 70% training, 15% validation, 15% testing. As a result of this, we got:

- The training data length is 37834.
- The validation data length is 8108.
- The testing data length is 8107.

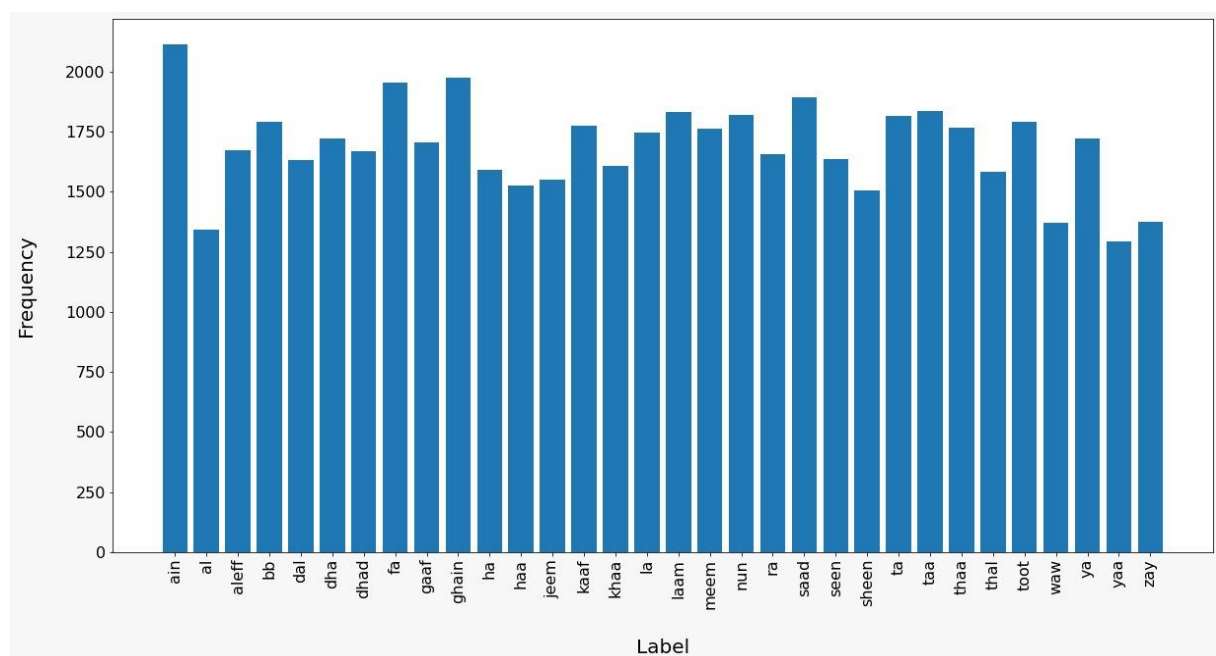

**Figure 3-7: ArASL2018 Dataset Histogram**

## 3.3 Define the Model

We trained various models to classify the letters. Many models can perform better than others, and the problems that existed in a model can solve in another model. Our case is considered a complex problem, so the traditional approaches cannot achieve our target.

### 3.3.1 Multilayer Model (ANN)

This model consists of one input layer, one or more hidden layers, and one final layer called the output layer. The number of neurons at the input layer equals the number of data input; in our case, it equals the number of pixels $64 \times 64 = 4096$. Hidden layers are chosen according to the complexity of data, and every hidden layer is responsible for the extraction of specific features. The number of neurons at the output layer equals the number of classifications, in our case, equals 32.

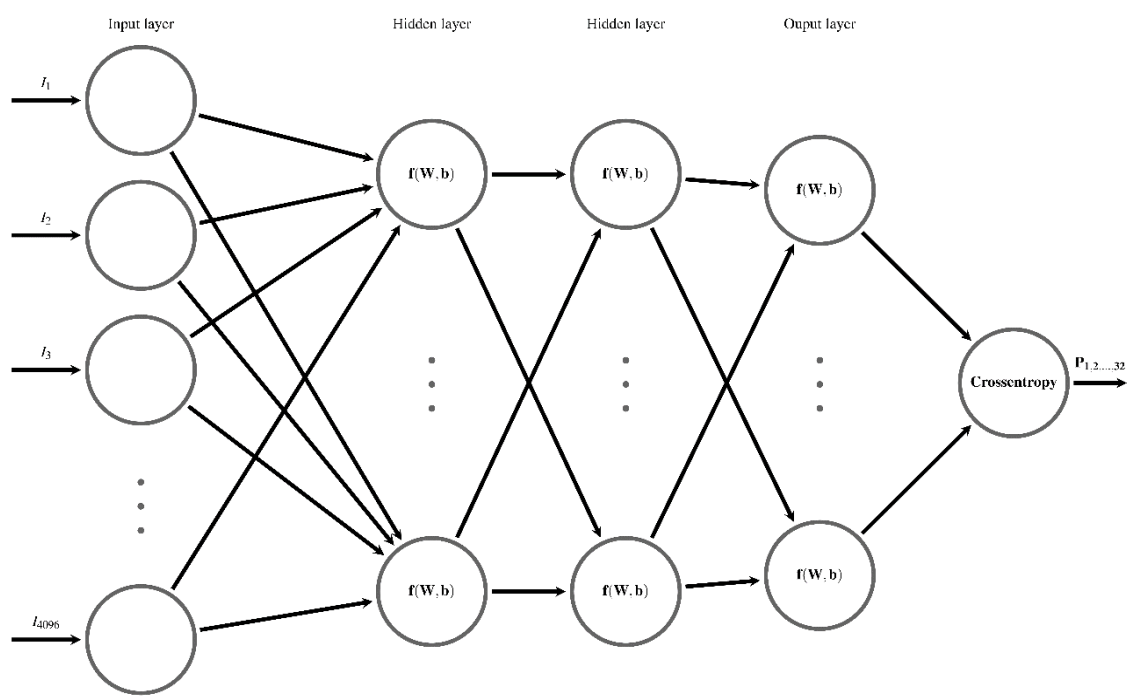

**Figure 3-8: Multilayer Algorithm**

The following details will explain the algorithm:

- By default, the algorithm deals with input data image by image, and we have more than 50 thousand images, so the training of the dataset will take much time. As a result, that, we divide the dataset into many batches to faster our model's training. And we used to stratify the function available in the Scikit-learn framework, which takes only the number of images from each class.

- Every neuron has a weight and bias that represent some data features. It must be to initialize them randomly. The manipulation of them increasingly or decreasingly can decide how the model's accuracy. The goal is to get better weights and biases, which achieve less error.

- Every batch will pass to the input layer, then pass to the first hidden layer, and calculate the neurons' output. After that, the next hidden layer will receive the previous layer's output and apply the same previous calculations, and so on until we get the last layer's output.

- The calculations are responsible for determining the performance of the model. They called activation functions like; step, sigmoid, Tanh, and ReLU. Some of them are used to easy models, and the others to complex ones.

- The output that we will get from the last hidden layer is free to be in any range. The softmax squeezes the output probabilities between 0 and 1. The architecture supposes that the higher probability is the correct output of the input.

- Next, the architecture measures the error using the loss function by comparing the obtained output with predictive output to estimate how much output corresponds with the predictive values, and the manipulation of the weights and biases will be done using optimizers like; SGD and Adam by propagating the error backwards from the output through the network.

**Activation Functions** [51]:

Activation functions are used to make the complicated mapping between the inputs and corresponding outputs by applying the function to the summation of input products with weights and biases and then feeding them to the next layer. The linear regression tasks can be dealt with by linear activation functions, but complicated tasks like image classification, speech recognition need non-linear functions to map between inputs and outputs. An activation function must be differentiable to implement the optimizer like backpropagation, stochastic gradient descent because it requires computing the errors for the gradient of weights. Figure 3-9 shows some activation functions and explains how their derivations seem.

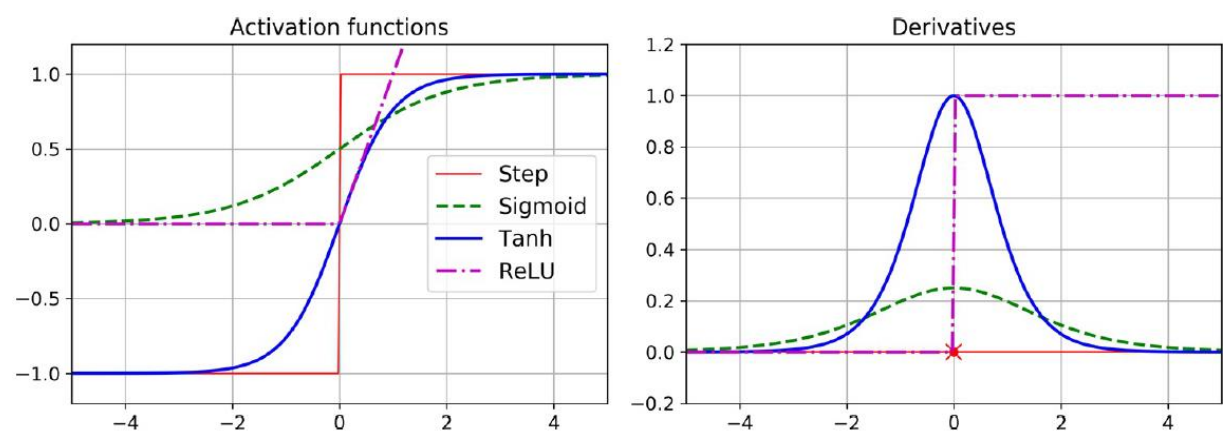

**Figure 3-9: Activation Functions and Their Derivatives** [28].

**a.  Step Function**

It is the simplest activation function and considered a hard limiter, activated when it has a value greater than a threshold. It cannot be used in multiclass classification tasks because it is limited between two values, $-1$ or $1$ . The equation that represents the function is:

$$f(x) = \begin{cases} 1, & x \geq 0 \\ -1, & x < 0 \end{cases} \qquad (3.1)$$

It is obvious from the equation that the derivative of the step function equals zero.

**b.  Sigmoid Function**

It is a non-linear function that bounds the input between 0 to 1. The equations that represent the function and its derivative are:

$$\sigma(x) = \frac{1}{1 + e^{-x}}$$

$$\acute{\sigma}(x) = \sigma(x)\big(1 - \sigma(x)\big) \qquad (3.2)$$

The function has a nonzero derivative everywhere, allowing the optimizer to update the weights and biases.

### c. Hyperbolic Tangent Function

It is like a sigmoid function, and they have S-shaped and continuously differentiable, but its values bound the input between $-1$ to 1. The equations that represent the function and its derivative are:

$$f(x) = 2\sigma(x)(2x) - 1$$

$$\acute{f}(x) = \frac{2}{1 + e^{-2x}} - 1$$

(3.3)

Compared to the sigmoid function, the gradient of tanh function is steeper, and its derivative is centred around zero, which often helps speed up convergence.

### d. ReLU Function

ReLU stands for a rectified linear unit and continuously differentiable except at zero, where the slope changes abruptly, causing a bouncing around the optimal values. However, it works very well and can be fast to compute because it is linear, and its derivative is constant. Thereby the calculations will be simple. The equations that represent the function and its derivative are:

$$f(x) = \begin{cases} x, & x \geq 0 \\ 0, & x < 0 \end{cases}$$

$$\acute{f}(x) = \begin{cases} 1, & x > 0 \\ 0, & x < 0 \\ not\ defined, & x = 0 \end{cases}$$

(3.4)

## 3.3.2 CNN Model

Using multilayer architecture is not sufficient for image recognition tasks because if we get a simple image with $100 \times 100$ pixels, we will have 10000 neurons, then feeding them in a layer with 1000 neurons, we will obtain 10 million connections at the first hidden layer, it is considered a huge number of links and hard to deal with it. It is noticed that the multilayer model can deal appropriately with small images that have few features. CNN is a proper solution that uses a specific layer to minimize the connection to achieve the classification.

### 1. Convolution Layers

It is an essential component of the CNN architecture used for feature extraction. The neurons in the first convolution layer do not connect to every neuron in the input data, but neurons in one layer are connected to other neurons in their receptive field (specific patterns in small regions of the visual field). Each neuron in the second convolution layer links neurons located within a small rectangle in the first convolution layer. Figure 3-10 shows the connections between the layers. The first layer in the architecture is responsible for extracting some features, the next layer for other features, etc.

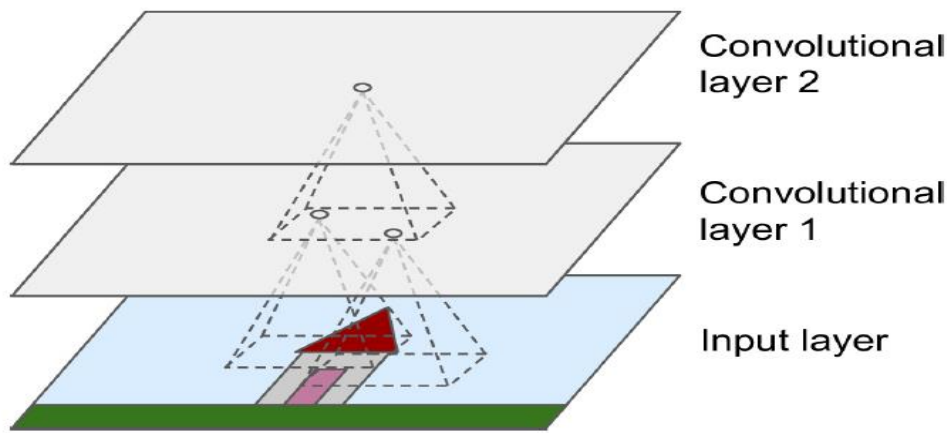

**Figure 3-10: CNN Layers with Rectangular Local Receptive Fields** [28].

The array that combines the neurons is called the kernel. The operation above does not guarantee each kernel's centre to overlap the input layer's outermost element. Padding, precisely Zero Padding, is a solution to avoid adding zeros around the inputs that can overlap the outer element of the input layer. Stride is "the distance between two consecutive Kernels" [29].

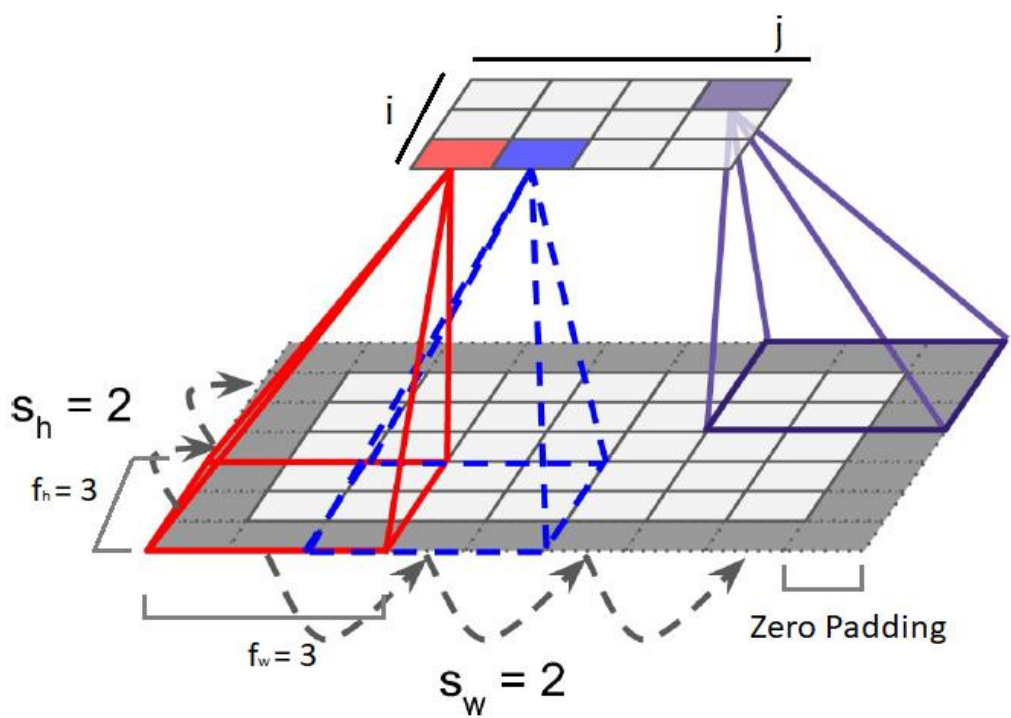

**Figure 3-11: Connections Between Layers, Adapted from**s [28].

### 2. Feature Maps

The convolutional layer can be represented in 3D which every layer has multiple feature map. A feature map is considered a filter that explores features like vertical lines and horizontal lines. Figure 3-12 shows convolutional layers with multiple feature maps.

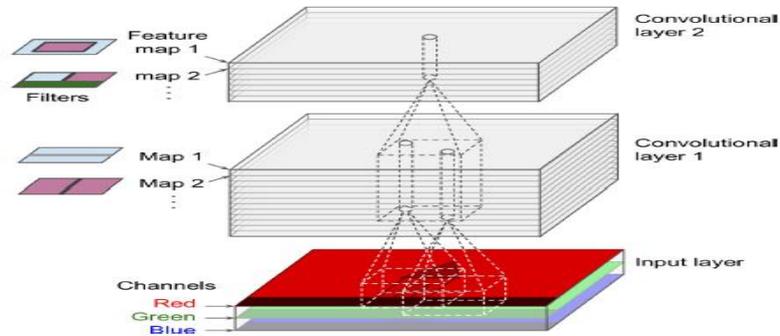

**Figure 3-12: Convolutional Layers with Multiple Feature Maps** [28].

The equation that shows how the convolutional layer computes the output of a given neuron is:

$$Z_{i,j,k} = b_k + \sum_{u=0}^{f_h-1} \sum_{v=0}^{f_w-1} \sum_{k'=0}^{f_{n'}-1} X_{i',j',k'} \cdot W_{u,v,k',k} \qquad with \begin{cases} i' = i \times s_h \times u \\ j' = j \times s_w \times w \end{cases} \qquad (3.5)$$

Where:

- $Z_{i,j,k}$: the neuron's output in row i, column j in feature map k of the convolutional layer.
- $X_{i',j',k'}$: the output of the neuron in layer L − 1, row i′, column j′.
- $b_k$ : the bias term in layer L.
- $W_{u,v,k',k}$ : the connection weights.

### 3.   Pooling Layers

This layer decreases the computations by shrinking the inputs. The pooling layer is like the convolutional layer, connected partially to the previous layer's outputs, located with the small rectangle receptive field. There are two forms of pooling: Max pooling and Mean pooling. Max pooling is the most popular form, which takes the maximum value in the higher-level feature layer. Mean pooling takes the average of all the elements in the higher-level feature layer. Max and Mean pooling are shown in Figure 3-13.

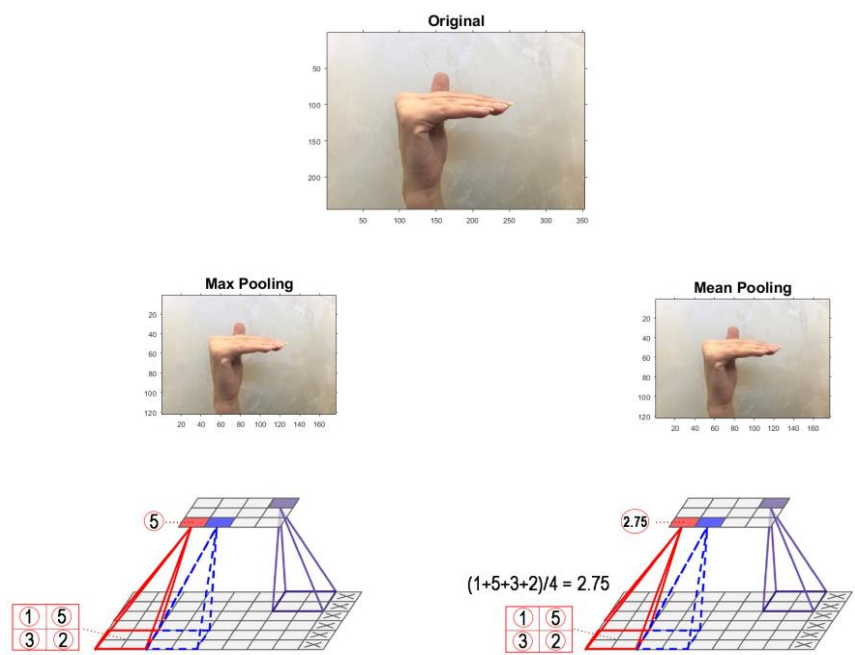

**Figure 3-13: Max Pooling and Mean Pooling with (2×2 Pooling Kernel, Stride 2, Zero Padding)**

It is noticed from the above figure that the pooling layer with 2×2 pooling kernel and stride 2 decreases the image to a quarter of the original image. This stage will reduce the computations, memory usage, and the number of parameters, thereby easing the extraction of features from the image.

### 4.   Dropout Layer

A huge number of parameters in the network gives it the flexibility to tend to overfit. One solution to avoid overfitting is using the early stopping technique, where the network stores the parameters at the best values when the validation set worsens. With unlimited computations, the early stopping technique will be aggressive and consume more time so, the best way to avoid overfitting is using the regularizes. Dropout is one of the most regularization techniques. The term dropout refers to dropping out the neuron and incoming and outcoming connections temporarily from the network, as shown in Figure 3-14.

The neurons that will be dropped out are chosen randomly, or simply, each neuron in the training set will be multiplied with a factor called dropout rate ρ. The dropout rate can be selected from between 0.4-0.5 in CNN. In the testing set, the neurons will not be dropped out. For more explanation, suppose $P$=0.5, the neurons during testing will be connected twice more than neurons in training. So, each neuron in testing has a total input signal twice larger than what each neuron in training has, and the performance will not be ok in this case. Each neuron input's signal in testing will be multiplied by 0.5 to compensate for the difference in neurons before and after training, as shown in Figure 3-15.

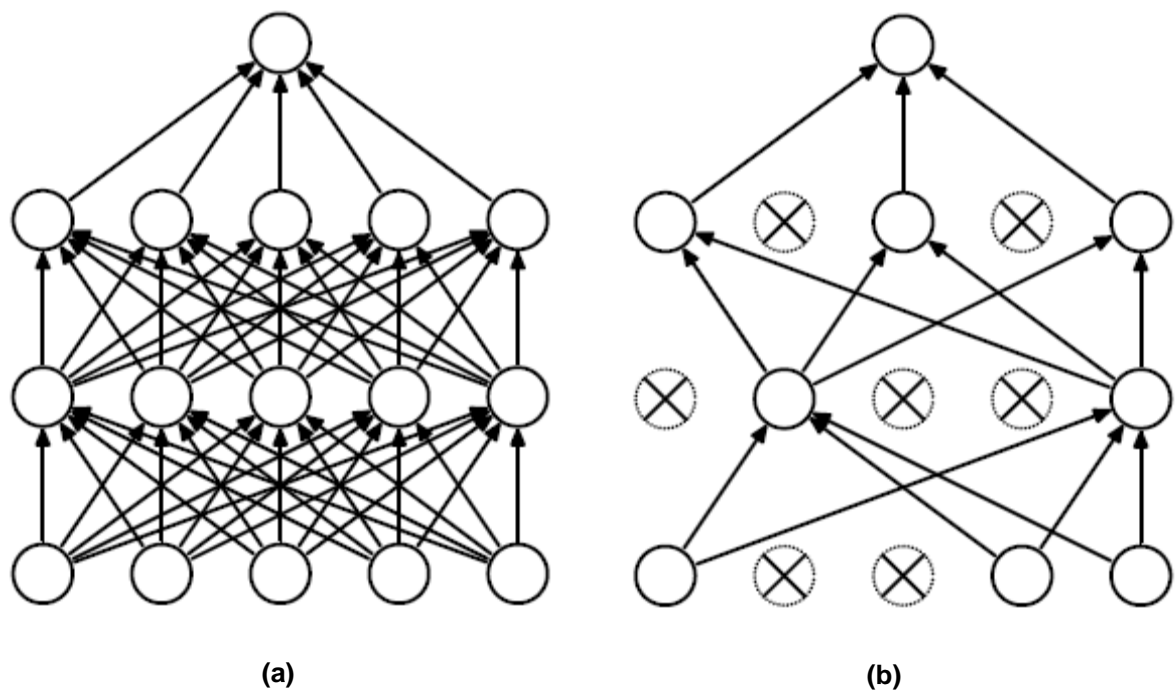

**(a)**                      **(b)**

**Figure 3-14: (a) Network Without Dropout, (b) Network With Dropout.** [52]

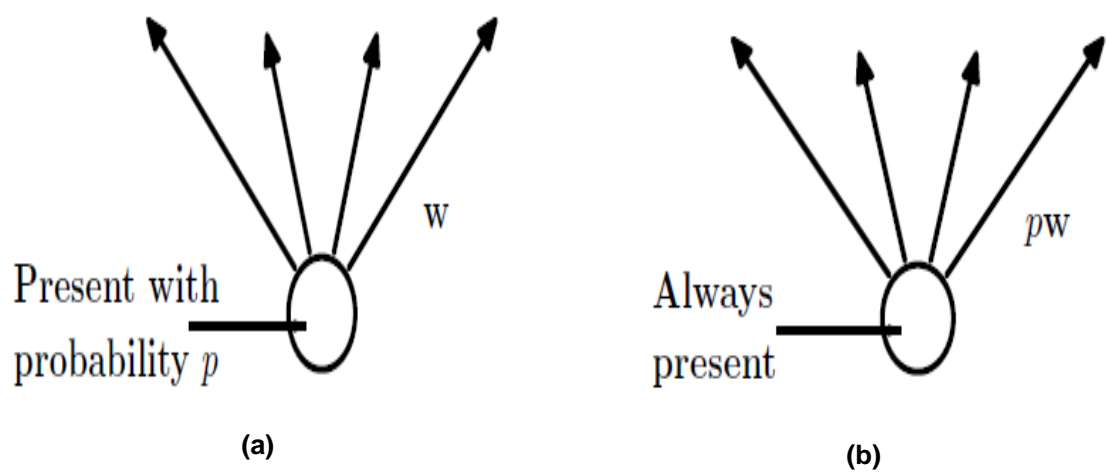

**(a)**                      **(b)**

**Figure 3-15: (a) Neuron at training, (b) Neuron at testing.** [52]

### 5. Fully Connected Layer

This layer transforms the last convolutional layer into a one-dimensional array and connects to one or more dense layers, in addition to a dropout layer after each dense layer, with a 0.5 dropout rate will reduce overfitting. A non-linear activation function follows the final fully connected layer to estimate inputs classification according to the output probabilities.

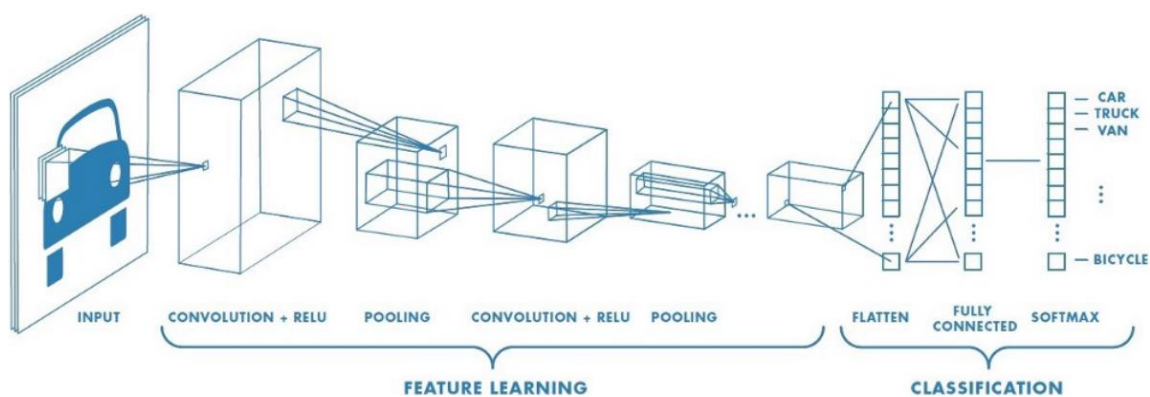

**Figure 3-16: CNN Architecture** [53]**.**

Figure 3-16 shows the general architecture of CNN. It is noticed that there are convolutional layers followed by a ReLU function or others, then another pooling layer, and so on. The previous steps are considered feature layers and make the image smaller and smaller through the architecture until it reaches the classification layers. Stages do the classification: flattening the last layer to pass it into a fully connected layer and then passing the fully connected layer into softmax function to classify the images according to the estimated probabilities.

## 3.3.3 ResNet-18

Many challenges face the building of CNN models, like specifying the number of layers and their size, initialization of weights, and biases. In bad initialization, the model can consume a lot of time to complete the task. Many architectures have been developed over the years, and they have a good impact on improving the trained models. On the other hand, these architectures simplify dealing with data without deep knowledge in this field. One of them is ResNet-18 architecture.

ResNet-18 is using a skip connection signal to train the model. A skip connection is a signal that passes into the layer in addition added to the output layer. In a usual architecture, the goal is training the $f(x)$ but in the residual training the architecture will be forced to train $f(x) + x$. Figure 3-17 shows the residual training.

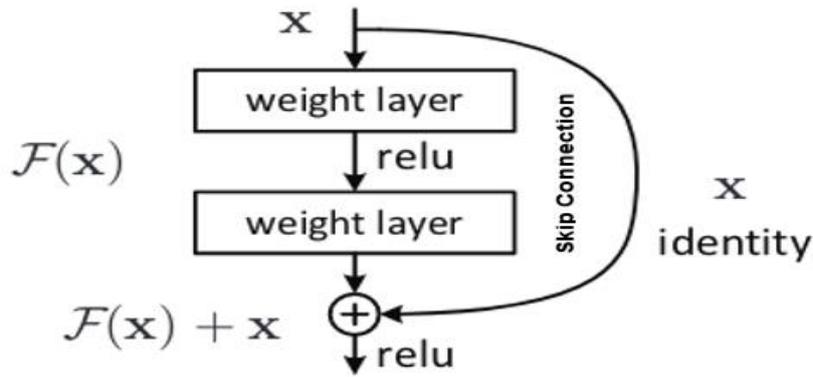

**Figure 3-17: Residual Learning**

The initial weights are usually close to zero so, the outputs follow them with values close to zero. In the case of adding the skip connection, the outputs will be clone from the inputs. This will accelerate the training faster than without the skip connection because the network progresses before the layers start learning.

ResNet-18 has 18 layers where the convolution layers and fully connected layer are just counted. Let us analyse ResNet's architecture that is shown in Figure 3-18. The input image is passed to a convolutional layer with a 7×7 kernel, 64 feature maps stride 2, zero Padding. Then is fed to a max-pooling layer with a 7×7 kernel, stride 2, zero Padding. After them, there are four identical convolutional networks. Every ConvNet has two residual units, and each residual unit consists of two convolution layers with a 3×3 kernel, stride 2, zero Padding. It is noticed that the feature maps are doubled every identical ConvNets, and the convolution layers' size is minimized to half in height and width. Next, the last layer is fed to the average pooling layer, then a fully connected layer and softmax to estimate the probabilities.

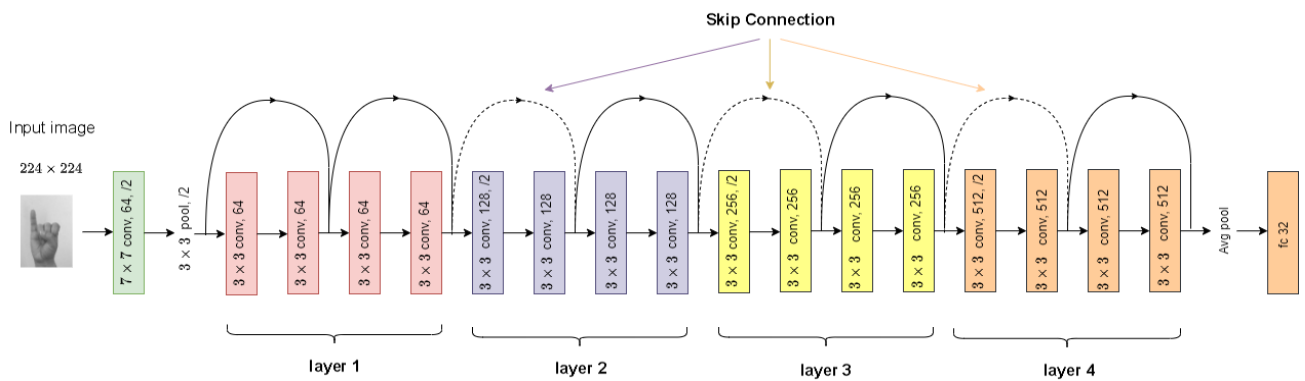

**Figure 3-18: ResNet-18 Architecture, Adapted From** [54].

## 3.4 Define A Loss Function and Optimizer

To assist the model, we need a loss function to ensure that the Training is doing well. And we need an optimizer to update the weights inside the network.

A loss function is considered one of the pillars when training the model in image classification. However, this loss function assists the learning for the model over the training dataset using the weights and biases through the network. For example, it can be calculated by taking the difference between the predicted and actual classes.

There are two common categories of loss function like **regression** loss functions and **classification** loss functions. However, regression seeks to find a predicted continuous value depending on many parameters in the model, while classification chooses an output from a set of categories. Here are many examples of regression and classification losses that are commonly used in these problems.

**Regression Loss Functions:**

- **Mean Square Error:**
  It is considered a performance technique that determines how much error the model obtains comparing with its predictions. The equation is formed by taking the difference between prediction from the model and the actual, then averaging these differences to get the total magnitude error.

$$MSE = \frac{1}{n} \sum_{i=1}^{n} (y_i - \hat{y}_i) \tag{3.6}$$

Where the symbols represent the following:

**MSE**: Mean Squared Error.

**n**: total number of predictions/actual data.

$y_i$: actual value.

$\hat{y}_i$: predicted value.

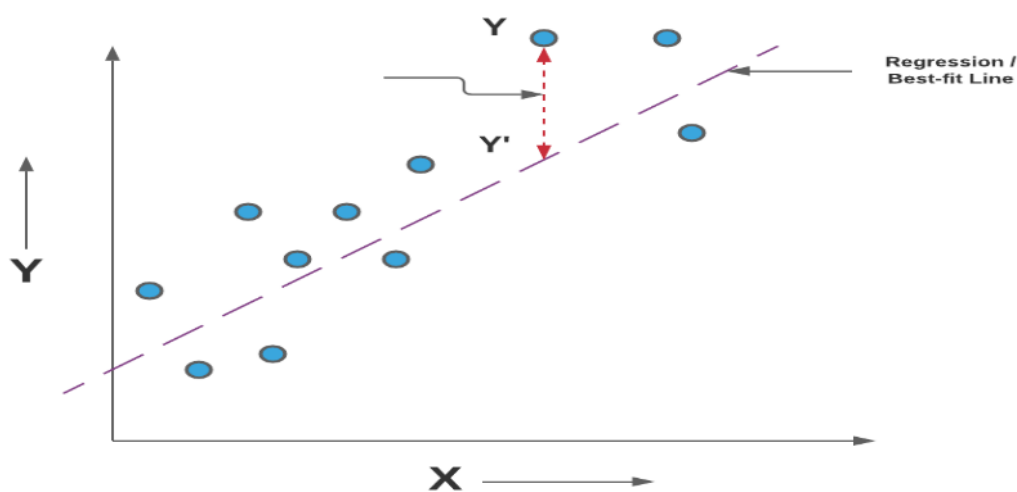

**Figure 3-19: Mean Square Error**

- **Mean Absolute Error:**

It is almost like MSE. Using one of them depends on the data distribution. When we have large outlier districts, MSE will neutralize the negative and positive outliers, giving a wrong prediction. It is preferred to take the absolute values. The equation is formed by taking the sum of the absolute differences between the prediction and the actual values, then obtain the average.

$$MAE = \frac{1}{n} \sum_{j=1}^{n} |y_i - \hat{y}_i| \qquad (3.7)$$

Where the symbols represent the following:

**MAE:** Mean Absolute Error.

**n**: total number of predictions/actual data.

$y_i$: actual value.

$\hat{y}_i$: predicted value.

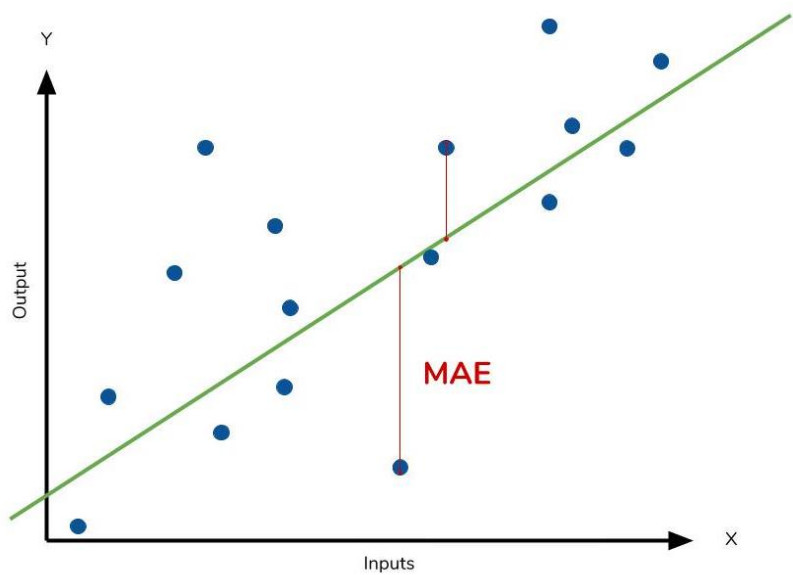

**Figure 3-20: Mean Absolute Error**

**Classification Loss Functions:**

- **Mean Squared Error:**

  Taking the difference between prediction from the model and the actual, then averaging these differences to get the total magnitude error.

- **Cross-Entropy Loss (Log Loss):**

  It is often used in classification problems. The output probability from a sigmoid or a softmax then enters the cross-entropy function that assesses how much this model classifies well.

  It has two forms, binary cross-entropy, and multi-class cross-entropy. The following equations clarify each of them.

- **Binary Cross-Entropy:**

$$BCE = y * \log(p) + (1 - y) * \log(1 - p) \tag{3.8}$$

P = Prop(y=1), output from a sigmoid activation binary class label y.

The following graph illustrates the loss vs. predicted probability for a binary classifier for each y=1 and y=0.

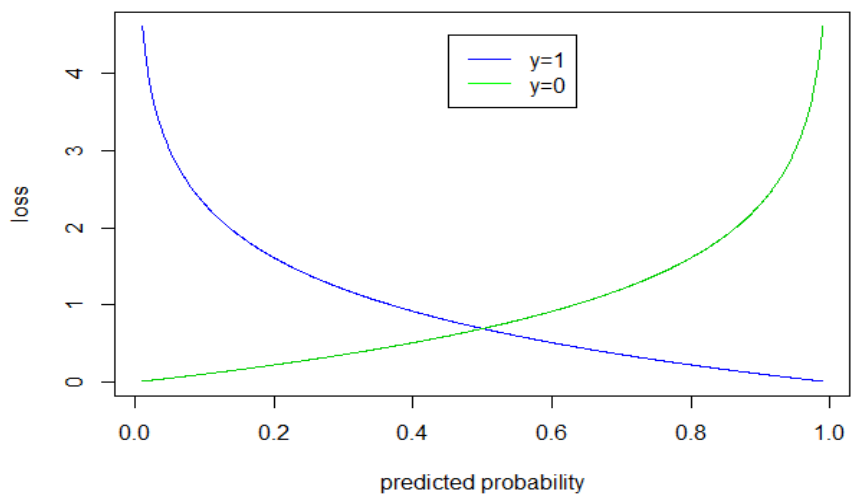

**Figure 3-21: Loss (BCE) vs. Predicted Probability**

- **Categorical Cross-Entropy:**

$$CCE = -\frac{1}{m}\sum_{i=1}^{m}(\, y_i * \log(\hat{y}_i) + (1 - y_i) * \log(1 - \hat{y}_i)\,)$$

(3.9)

m: number of classes that are represented with one-hot encoding.

$y_i$: i[th] target class.

$\hat{y}_i$: the predicted probability of that input belongs to the i[th] class, computed with a softmax activation.

- **Optimizer:**

It is an algorithm or method used to tune the model's parameter (e.g., weights, biases, etc.) during the training to reduce the losses. Also, it affects the results from the model.

Many optimizers can be used with the models during the training. However, every optimizer has its advantages and disadvantages.

- **Gradient Descent:**

It is considered one of the most basic optimization algorithms, and it is a first-order optimization algorithm dependent on the first order of a loss function. It can be used in classification and linear regression problems. Also, it is used in backpropagation algorithm. The following equation can express it:

$$\theta = \theta - \alpha \, \nabla J(\theta) \qquad\qquad (3.10)$$

$\boldsymbol{\theta}$: is the weight of the model.

$\boldsymbol{\alpha}$: is the learning rate.

$\boldsymbol{\nabla J(\theta)}$: is the derivative of the objective (loss) function for the weights.

- **Stochastic Gradient Descent (SGD):**

  The difference between this algorithm and basic gradient descent is that this algorithm updates the model's parameter for each training example in the dataset.

$$\theta = \theta - \alpha \, \nabla J(\theta;\, x(i);\, y(i))$$

$$(3.11)$$

Where:

$\{\,\boldsymbol{x(i)}, \boldsymbol{y(i)}\}$: are the training examples.

$\boldsymbol{\theta}$: is the weight of the model.

$\boldsymbol{\alpha}$: is the learning rate.

$\boldsymbol{\nabla J(\theta;\, x(i);\, y(i))}$: is the derivative of the objective (loss) function for the weights for every training example.

In this algorithm, the loss function has a lot of fluctuations and variance over the training.

- **Mini-Batch Gradient Descent**

  It combines the advantages of SGD and standard gradient descent. After every batch of the training dataset the parameters are updated.

$$\boldsymbol{\theta = \theta - \alpha \, \nabla J(\theta;\, B(i))} \qquad\qquad (3.12)$$

Where:

$\{\boldsymbol{B(i)}\}$: the batches of training examples.

$\boldsymbol{\theta}$: the weight of the model.

$\boldsymbol{\alpha}$: the learning rate.

$\boldsymbol{\nabla J(\theta;\, B(i))}$: objective (loss) function derivative for the weights for every batch.

Learning rate: The gradient tells us where the function has the highest rate of change, but it does not tell us the value of steps to reach the optimal value. Small steps lead the algorithm to the best solution but slow progress. Large steps are fast but lead the algorithm to bounce around the optimal values.

- **Adam Optimizer:**

Adam is derived from adaptive moment estimation. It takes the advantages of both AdaGrad and RMSProp algorithms to provide an optimization that can deal with sparse gradient on noisy problems. It decays exponentially average of past gradients so, converging to the optimal solution fast. **Table 3-1** shows a brief comparison between SGD and ADAM.

**Table 3-1: Comparison between SGD and ADAM**

| Optimizer | Advantages | Disadvantages |
|-----------|------------|---------------|
| SGD | • Can best fit and generalize the dataset after extensive training | • Cannot deal with global minima.<br>• It can be affected by choice of learning rate. |
| Adam | • Can deal with sparse gradient on noisy problems<br>• Can decay the learning rate through the learning | • Cannot generalize dataset perfectly |

# Chapter 4 DESIGN TESTING AND RESULTS

## 4.1 Model Training and Validation

### 4.1.1 Model Training

Model training means applying an algorithm to update the model parameters that best fit training data and predict the new data well. Models differ from others in the ability of data fitting. Each of them is suitable for a specific type of data and performs depending on the complexity of data. In our project, the data was trained using; ANN, CNN, ResNet-18.

The training performance depends on several factors, and without them, the training will be invalid and absorb more time. The factors that can improve the training are:

- **Model Architecture:** Before building the model, you should know the nature of the data. In our case for image classification, the deep neural network does not guarantee to extract the image features and achieve high accuracy, so choosing a powerful architecture like CNN or transfer learning is preferred.
- **Data Plenty:** Machine learning models need many data to fit them properly. The data that we ran was adequate to train the model. In case of lack of data, Data augmentation can be used or using transfer learning to compensate it.
- **The Optimizer:** The weights will be initiated randomly, so the optimizer will update the weights to reach the minimized error. We used ADAM and SGD to update the weights in the models.
- **The Number of Epochs:** The training of data needs several cycles to reach the proper parameters. There are two approaches to choose the number of epochs; specifying the number of epochs directly or using the early stopping technique since the training will stop when there is no progress. We set the number of epochs to 20 for the models.

### 4.1.2 Model Validation

It means how the model will estimate performance after the training and is used essentially to avoid overfitting. Overfitting occurs when the model performs well, generalizes the data quite on the training but performs worse on the validation. The validation is applied using evaluating the validation error every training stage to find the minimum error, then stopping the training and saving the parameters.

### 4.1.3 Graphs: Training, Validation Accuracy and Loss

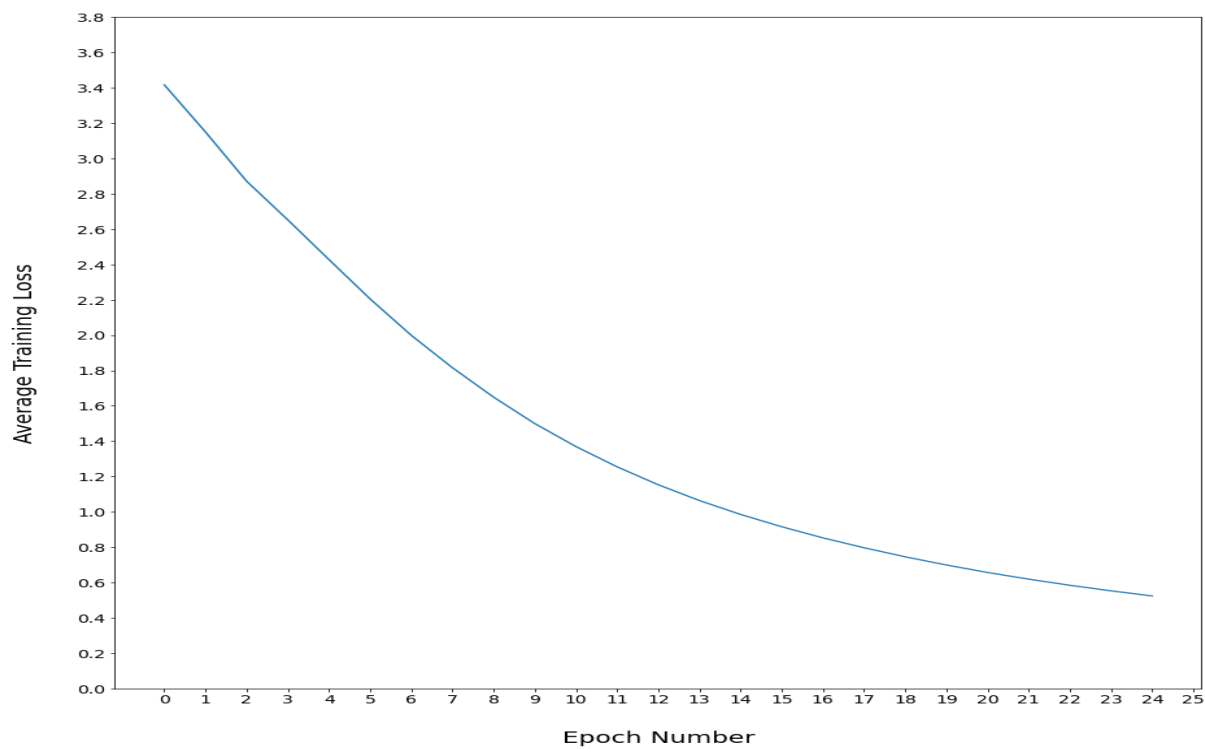

**Figure 4-1: Progress of Average Training Loss of ANN with SGD, lr=0.1 Through the Epochs.**

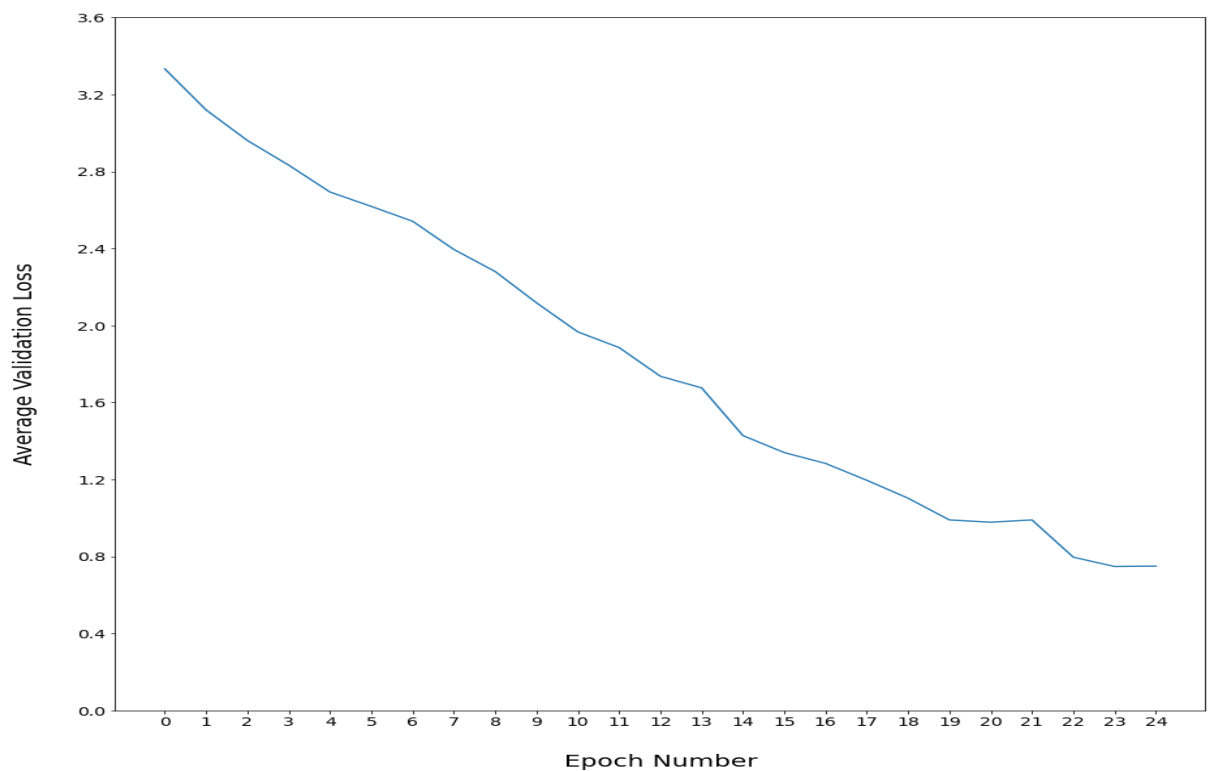

**Figure 4-2: Progress of Average Validation Loss of ANN with SGD, lr=0.1 Through the Epochs.**

Figure 4-1 shows that the average loss of accuracy training for an ANN model using SGD optimizer and lr = 0.1 decreases slowly with increasing number of epochs. Also, Figure 4-2 shows that the progress of validation loss for an ANN model does not decrease smoothly with increasing number of epochs. Finally, we can conclude that ANN needs a lot of time to be trained and to extract the features from the images.

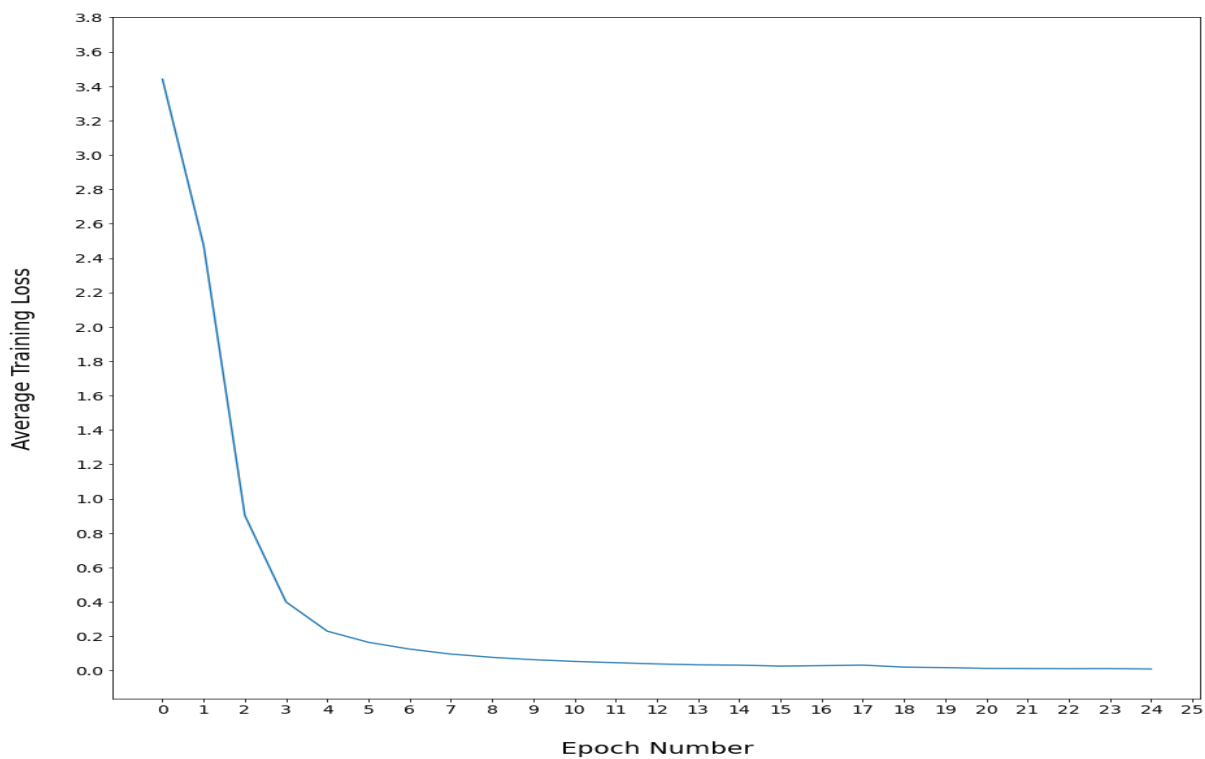

**Figure 4-3: Progress of Average Training Loss of CNN with SGD, lr=0.1 Through the Epochs.**

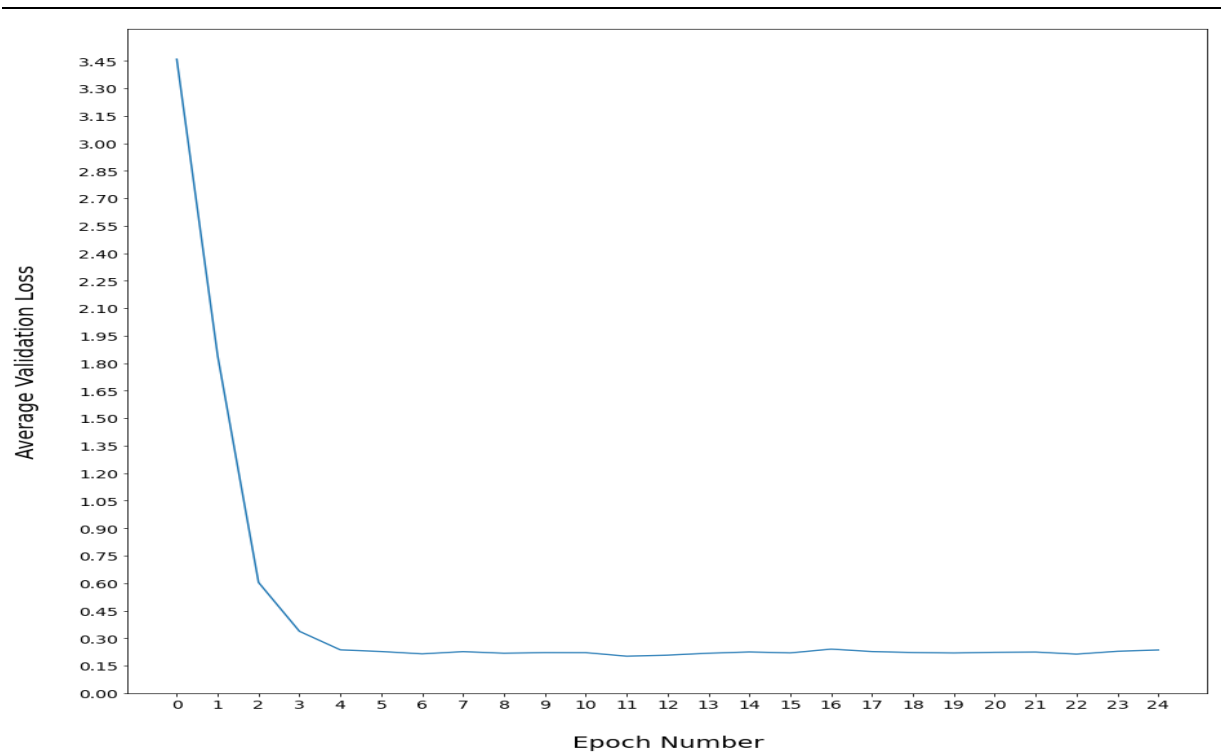

**Figure 4-4: Progress of Average Validation Loss of CNN with SGD, lr=0.1 Through the Epochs.**

CNN performs much better than ANN, Figure 4-3 and Figure 4-4 show that training loss and validation loss decrease rapidly through the first 5 epochs and there is not obvious decreasing until epoch number 24. Also, it is noted that CNN training loss and validation loss approaches to zero.

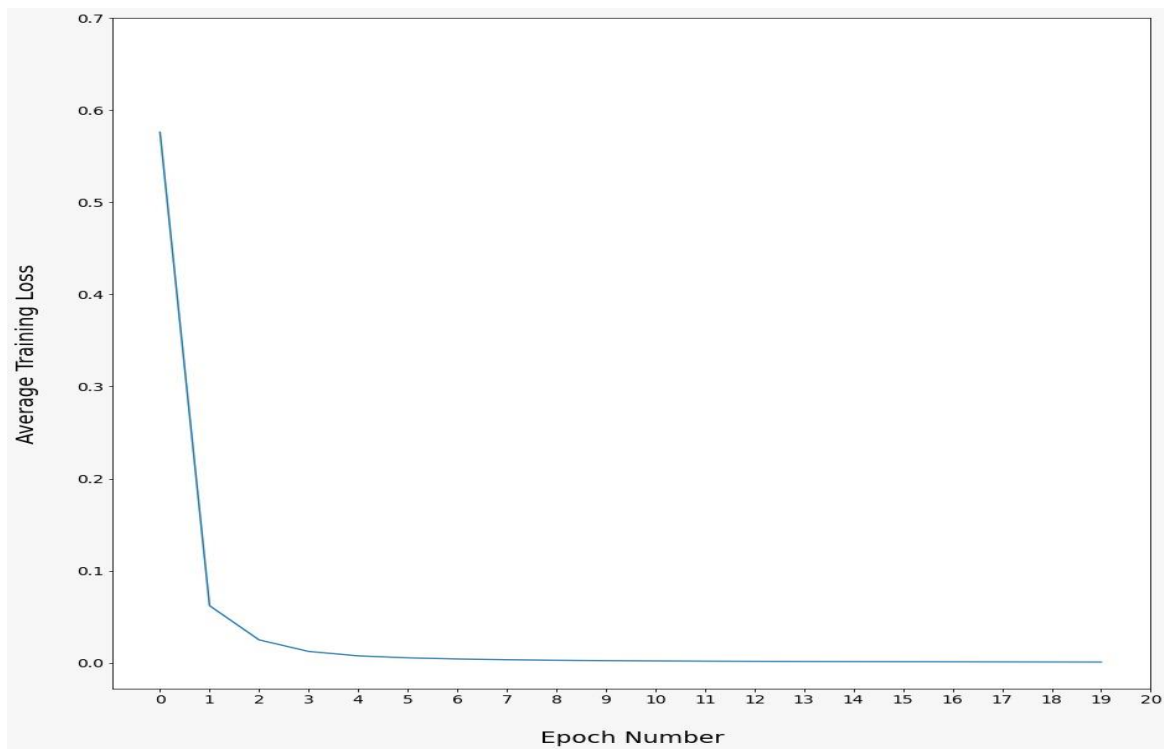

**Figure 4-5: Progress of Average Training Loss of ResNet-18 with SGD, lr=0.1 Through the Epochs.**

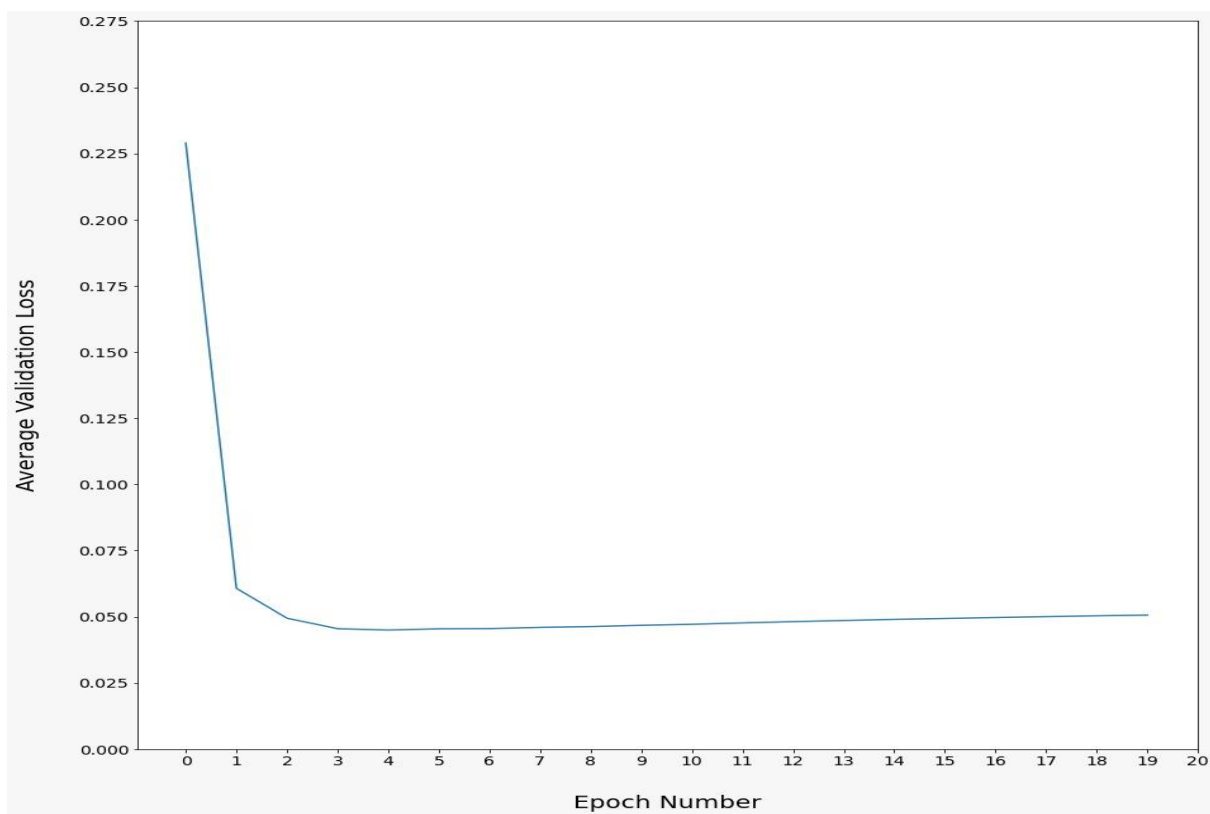

**Figure 4-6: Progress of Average Validation Loss of ResNet-18 with SGD, lr=0.1 Through the Epochs.**

Figure 4-5 and Figure 4-6 show that ResNet-18 model performs better than ANN and CNN. It is noted that it starts with low loss and it decreases through the first 3-4 epochs. However, the training loss decreases slightly and stays at low average loss, but average validation loss increases slightly after epoch 4 which means the model begins overfit over the data. However, we save the model at the least validation loss to avoid overfitting.

## 4.2 Testing and Results

After training the model, we need to validate it using a testing dataset. The testing stage performs the model on a dataset that the model did not see it. The image will pass into the model as an input, and the output will be one of 32 classes. If a matching between the true output and predicted output occurs, it contributes rising of model accuracy. The models' performance and their average accuracies are shown in Table 4-1.

**Table 4-1: Comparison Between Different Models' Accuracies**

| Model | Optimizer | Learning rate | Average Test Accuracy |
|---|---|---|---|
| ANN | SGD | 0.01 | 77.78 % |
| | SGD | 0.10 | 27.50 % |
| CNN | SGD | 0.01 | 93.00 % |
| | SGD | 0.10 | 95.80 % |
| Transfer Learning (ResNet-18) | SGD | 0.01 | 99.21 % |
| | SGD | 0.10 | 99.36 % |
| | ADAM | 0.01 | 99.00 % |
| | ADAM | 0.10 | 96.76 % |

It is noticed that the ResNet-18 has higher accuracy than other architectures. The model corresponds to the predicted values with actual values at the testing stage, and Figure 4-7 shows how much the testing set corresponds to the actual values at ResNet-18.

In general, each image will pass into the model and distribute to 32 classes in different accuracy so that the higher accuracy will be the predicted value. The misleading will have occurred in several cases, like; the similarity between the images, in which the model usually cannot discern images that have similarity, and the unclear images, in which the model cannot extract the features correctly because of the noises in the image.

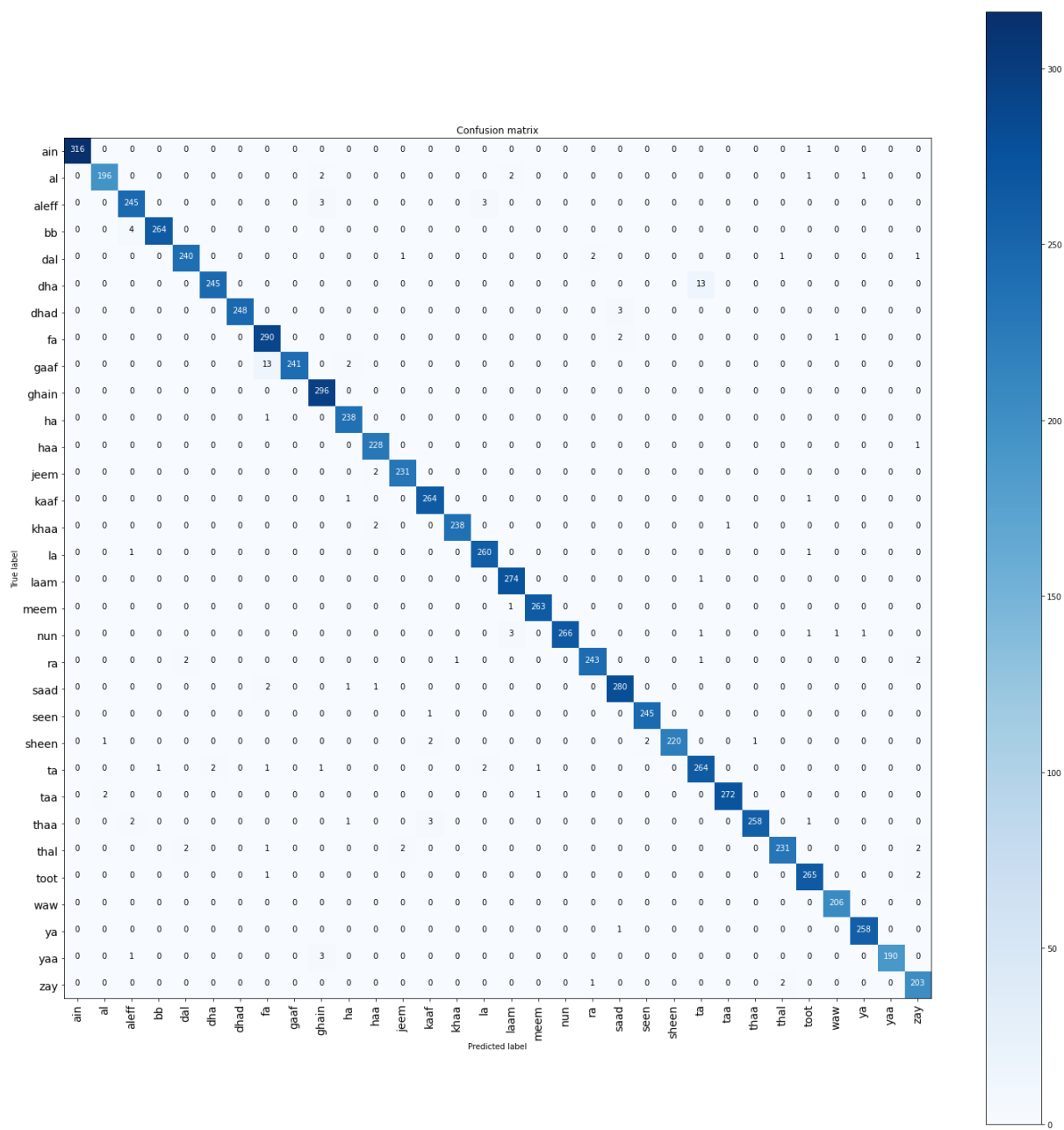

**Figure 4-7: Confusion Matrix of ResNet-18 Shows the Matching Between the Predicted Values with True Values at The Testing Stage.**

**Figure 4-8: Confusion Matrix of CNN Shows the Matching Between the Predicted Values with True Values at The Testing Stage.**

Figure 4-8 shows how much the testing set corresponds to the actual values at CNN. It is noticed from the distribution of predicted and true values that CNN performs well and the misleading between predicted and true values are considered little much.

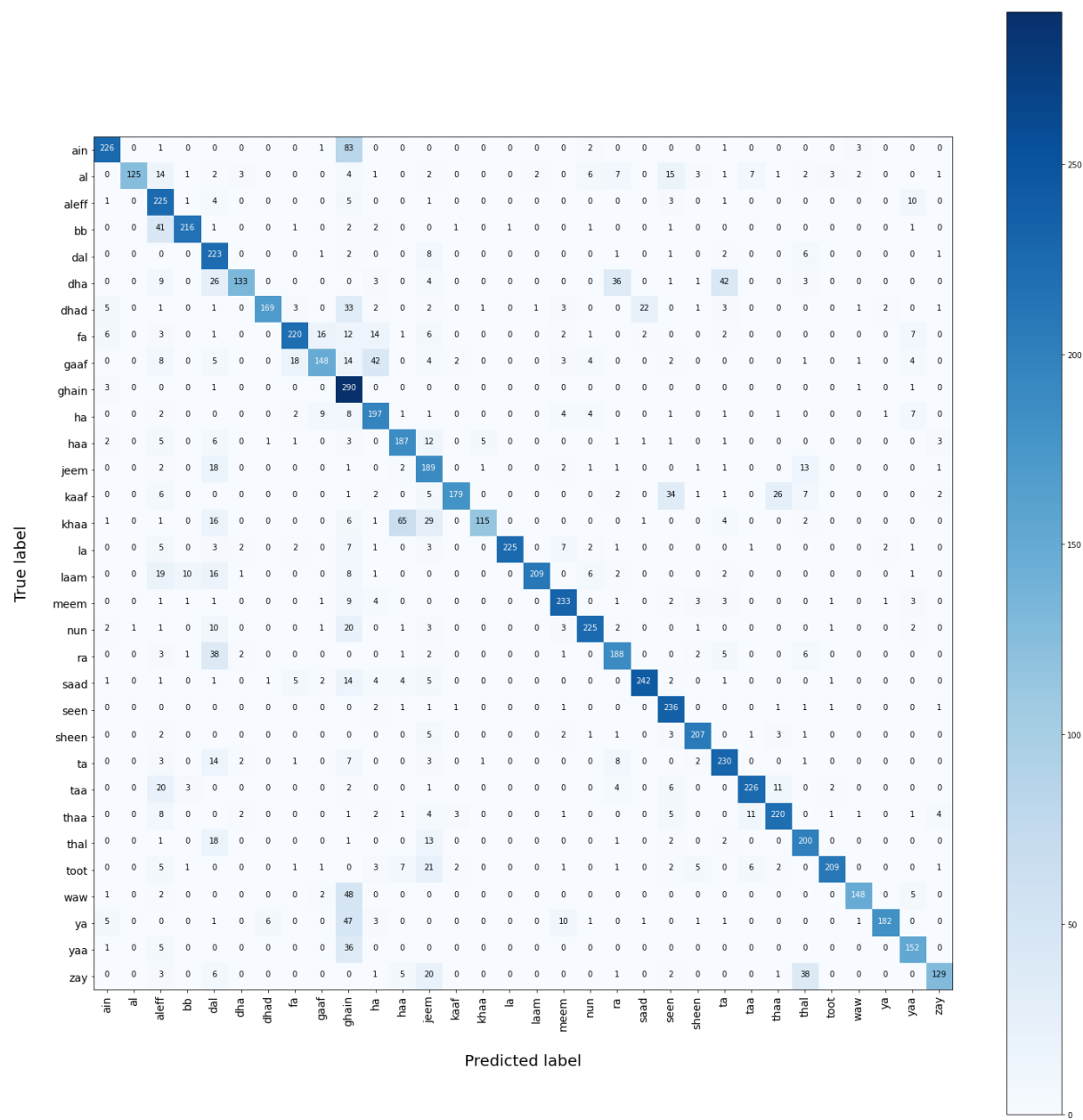

**Figure 4-9: Confusion Matrix of ANN Shows the Matching Between the Predicted Values with True Values at The Testing Stage.**

Figure 4-9 shows how much the testing set corresponds to the actual values at ANN. It is noticed from the distribution of predicted and true values that ANN does not perform well, and there is big misleading in some of the classes, indicating that ANN is not a valid image classification.

## 4.3 Model Inferencing

It is a process of feeding a new input image (unseen image) into a trained DNN model. Figure 4-10 shows the Flowchart of model inferencing.

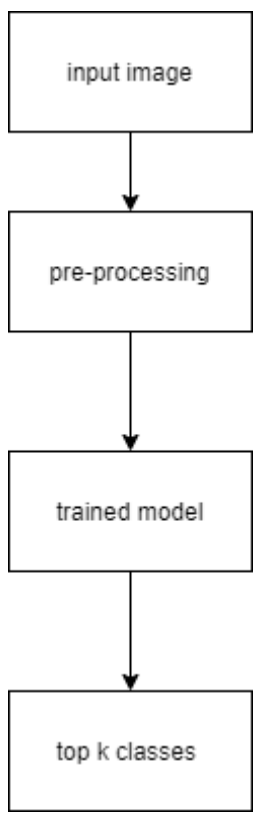

**Figure 4-10: Inferencing Flowchart**

We tried to examine the model performance by inserting an image shown in Figure 4-11 (a). The input image will pass into the pre-processing stage, which does some operations, including resizing, centre crop, converting to a grayscale image, etc. The image after pre-processing is shown in Figure 4-11 (b). The pre-processed image will pass into the trained model to classify the image to the predicted class. Figure 4-12 shows the prediction of the input image, and it is obvious that the model predicts the input correctly.

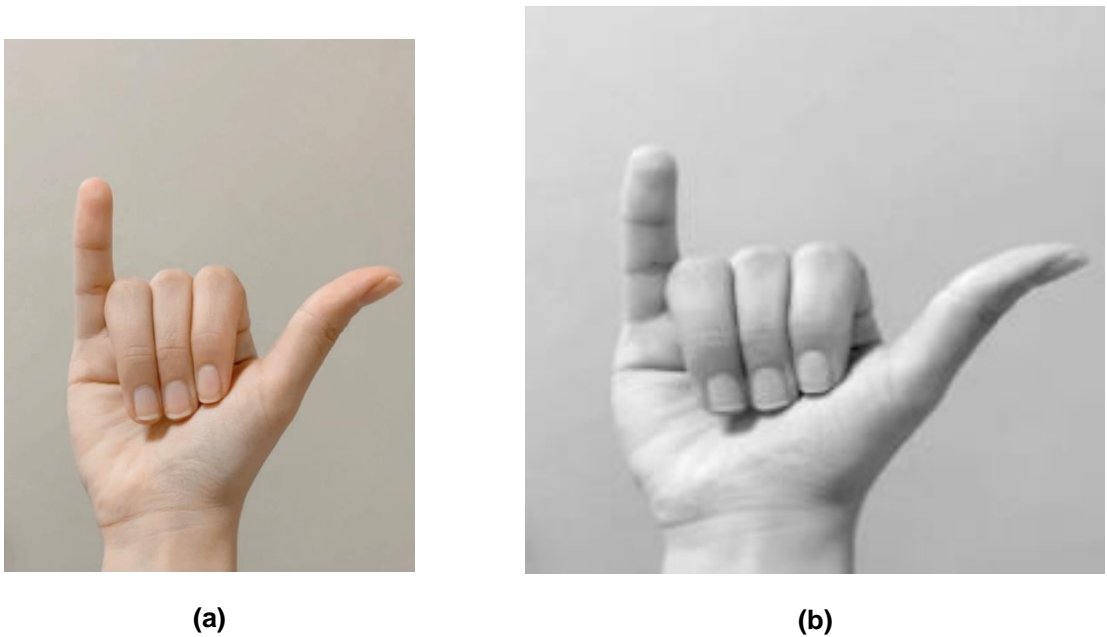

(a)                                    (b)

**Figure 4-11: (a) Input Image (Y$\bar{a}$), (b) Pre-processed Input Image**

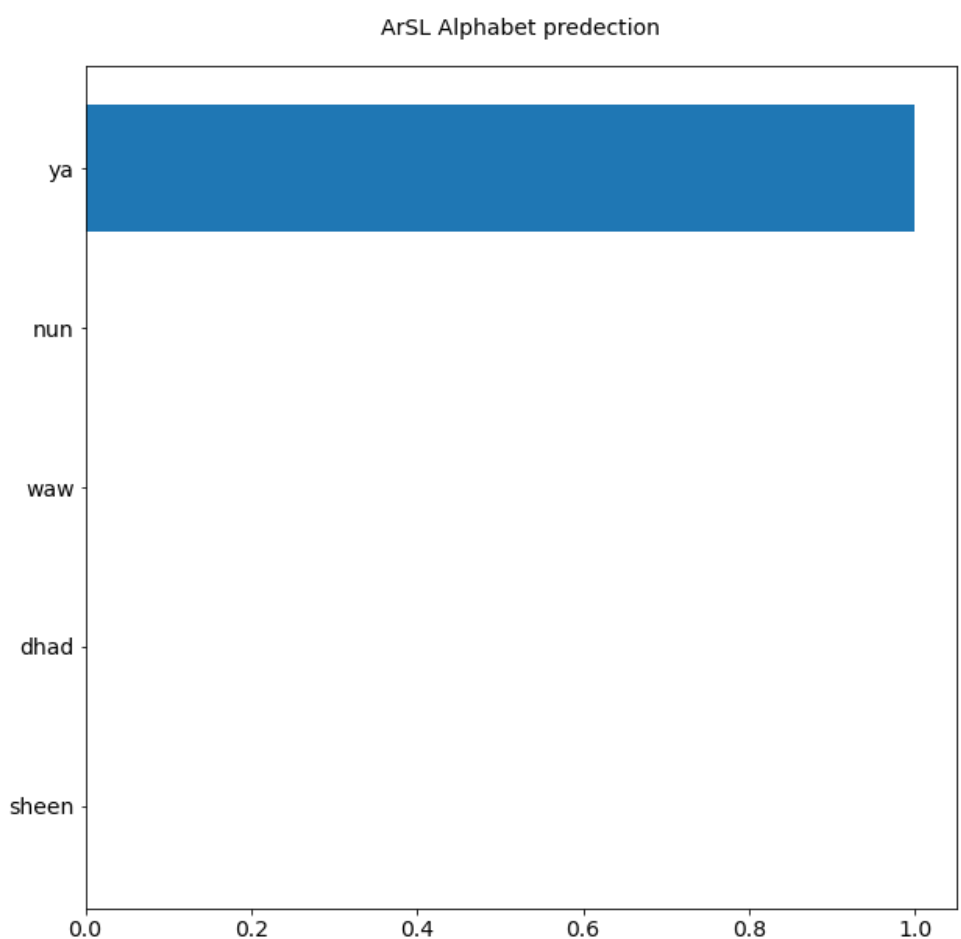

**Figure 4-12: ArSL Alphabet Prediction.**

## 4.4 New Collected Dataset: ArSLA-2021 Dataset

### 4.4.1 Overview

We are creating our dataset that is captured from many people of different ages, and it will be the first dataset of its kind for Arabic Sign Language Alphabets. Until now, we have collected about +10000 real-life images for 31 Arabic Sign Alphabets. Figure 4-13 shows from our dataset.

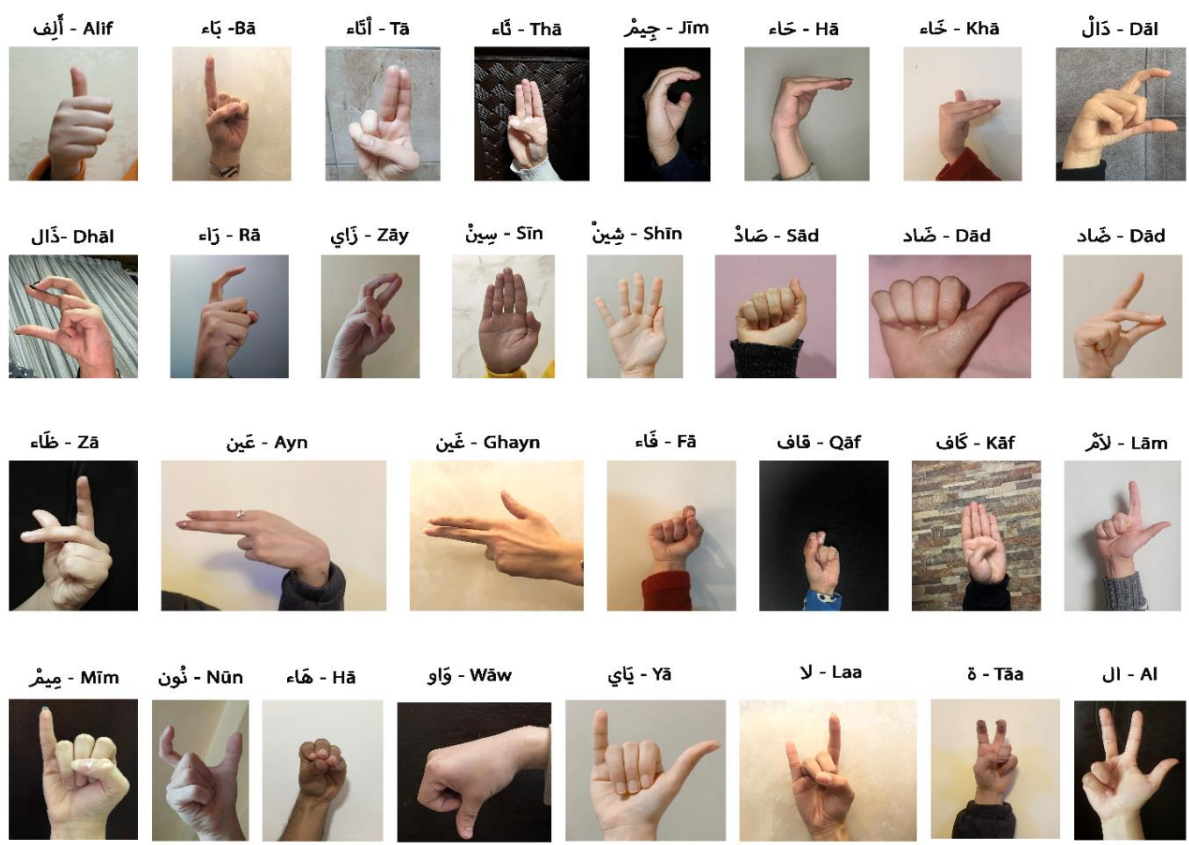

**Figure 4-13: Arabic Sign Language Alphabets Samples**

The dataset has been collected and captured by the people. Also, it is considered a real-life dataset with images captured under different conditions such as different light, background, image orientation, image size, image quality, etc.

**Table 4-2: Arabic Sign Language Alphabets, their numbers, and labels**

| # | Letter name in English Script | Letter name in Arabic script | # of Images | # | Letter name in English Script | Letter name in Arabic script | # of images |
|---|---|---|---|---|---|---|---|
| 1 | Alif | ا (ألف) | 395 | 17 | Zā | ظ (ظاء) | 329 |
| 2 | Bā | ب (باء) | 387 | 18 | Ayn | ع (عين) | 340 |
| 3 | Tā | ت (تاء) | 385 | 19 | Ghayn | غ (غين) | 337 |
| 4 | Thā | ث (ثاء) | 379 | 20 | Fā | ف (فاء) | 331 |
| 5 | Jīm | ج (جيم) | 388 | 21 | Qāf | ق (قاف) | 327 |
| 6 | Hā | ح (حاء) | 378 | 22 | Kāf | ك (كاف) | 332 |
| 7 | Khā | خ (خاء) | 347 | 23 | Lām | ل (لام) | 335 |
| 8 | Dāl | د (دال) | 343 | 24 | Mīm | م (ميم) | 328 |
| 9 | Dhāl | ذ (ذال) | 335 | 25 | Nūn | ن (نون) | 356 |
| 10 | Rā | ر (زاء) | 335 | 26 | Hā | ه (هاء) | 354 |
| 11 | Z āy | ز (زاي) | 340 | 27 | Wāw | و (واو) | 351 |
| 12 | Sīn | س (سين) | 340 | 28 | Yāa | ياء (ياء) | 346 |
| 13 | Shīn | ش (شين) | 353 | 29 | Tāa | ة (ة) | 344 |
| 14 | Sād | ص (صاد) | 340 | 30 | Al | ال (ال) | 341 |
| 15 | Dād | ض (ضاد) | 339 | 31 | Laa | لا (لا) | 339 |
| 16 | Tā | ط (طاء) | 334 | | | | |

This project is supported and consulted by Student Counselling Department at the University of Jordan, which has specialist interpreters for Arabic Sign Language. They help and consult us to achieve high-quality work in this field.

## 4.4.2 General Notes for ArSLA-2021 Dataset:

- It could be a benchmark for researchers in this field.
- It can be used for research and production.
- It will be shared as raw images. Everyone has the choice to do any processing on them.
- It is collected from more than 300 participants.
- Different resolutions have been got by different mobile phones.
- The dataset mainly consists of RGB images.
- These images are static.
- It will be annotated manually in the future.
- It will be made publicly available to support the Sign Language field.

## 4.5 System Limitations and Compliance with Design Constraints

### 4.5.1 System Limitations:

We faced many limitations while working on this project. The limitations are:

1. **Lack of Resources**

   We have just found one public dataset for ArSL alphabets that is publicly available, the ArASL2018 dataset. The available datasets have a limited set of conditions, including lighting, unique images, various background, etc.

2. **Hardware Limitations** [55]

- The training of deep learning models usually requires GPUs, with enough disk memory and RAM size that can accelerate the model training, but unfortunately, we do not have available GPUs at The University of Jordan.
- We have used the available GPUs that are offered by google colab. Unfortunately, google colab has some limitations:
  - The resources are not guaranteed.
  - The usage limits change depending on the availability.
  - Time limitations for continuous model running, which is at max 12 hours.
  - Memory limits to load the dataset into the model.
  - RAM limitation is used to perform the calculations.
  - GPUs' type is not guaranteed.

### 4.5.2 Design Constrains Compliance

The following points discuss the constraints that were put in place in the beginning, what we were able to solve, and what we could not:

1. **Availability of Data:**

We looked for any dataset that has multiple conditions that represent the real-life conditions of the data. We found one public dataset containing enough images, 54,049 images, that was trained on but with limitations to the number of unique participants, lighting, and complexity. However, we are collecting our dataset that overcome these issues.

2. **Computational Resources**

We have faced a challenge to train models using google colab due to restrictions on it. ArASL2018 dataset was trained using google colab without any problems and we have achieved an excellent performance. Unfortunately, we cannot using google collab to train models using ArSLA-2021 dataset.

3. **Response time (Inference Time):**

We calculated the inference time required to classify one image per model. It is noted from Table 4-3 that the induction time responds within a very short time.

**Table 4-3: Inference Time Results**

| Model | Optimizer | Learning Rate | Inference Time |
|---|---|---|---|
| ANN | SGD | 0.1 | 258 ms |
| CNN | SGD | 0.1 | 145 ms |
| Transfer Learning (ResNet-18) | SGD | 0.1 | 142 ms |

4. **Hyperparameter Choosing**

We have tested many hyperparameters depend on research papers that are considered relevant problem solutions.

5. **Knowledge and Experience in ArSL**

We have studied a lot of resources about alphabets in ArSL and we have met experts interpreters in ArSL to deliver a high quality of work.

## 4.6 Solution Impact

## 4.6.1 Societal Impact

Machine learning broadens our outlook on life. It makes a great leap in various fields like, industry, medicine, and social life. One of the machine learning branches is computer vision. Computer vision and its applications simplify the complicated tasks where some of these applications require much experience to extract the image's features.

Computer vision is not limited to technical issues but reaches toward humanity issues. In our project, computer vision is used to build a model that can recognize the Arabic sign language alphabets automatically. This step positively impacts the community, where it will spread awareness of sign language and ease the communication between deaf and normal people.

### 4.6.2 Economic Impact

Technology has a huge impact on the economy. If the Arabic sign language recognition model were deployed in smart devices, this would reduce the cost of owning special tools such as gloves, pressure sensors, and jump motion devices. The cost of hiring more interpreters will also decrease.

### 4.6.3 Environmental Impact

Our work is digital content, so when you use it there will be no waste. On the other hand, the use of solid materials will be a waste after corrosion, and this will have bad impacts on the environment.

### 4.6.4 Global Impact

If our model is deployed in smart devices, it will increase the communication between normal people and deaf people by interpreting the alphabets in ArSL and converting them into written Arabic text.

# Chapter 5 CONCLUSION AND FUTURE WORK

## 5.1 Conclusion

This project was designed to be the first step to help the deaf community by building a model that uses computer vision techniques to convert the ArSL alphabets into Arabic letters. Many machine learning techniques were used to build the model, and we chose transfer learning (ResNet-18) which achieved the highest accuracy.

## 5.2 Problems Faced

- It is a new field, so we have needed a lot of time to learn and grasp new concepts especially in deep learning and ArSL.
- We have not enough fund to buy GPUs to train new models using ArSLA-2021 dataset.
- Since we used the GPUs offered by google colab there was time limitation, and we had no control over the resources.
- COVID-19 restrictions have prevented us from collecting more data.

## 5.3 Recommendations for Future Work

We will expand the project to include:

- Publishing our dataset to be a starting point for someone else to continue the work.
- Training the model using our dataset and collecting more data.
- Collecting dataset for dynamic alphabets and collecting dataset to include words and continuous speech.
- Creating a model that will be used for real-time application.
- Deploying the model on the mobile application.